\documentclass[lettersize,journal]{IEEEtran}
\usepackage{amsmath,amsfonts}
\usepackage{algorithmic}
\usepackage{algorithm}
\usepackage{array}
\usepackage[caption=false,font=normalsize,labelfont=sf,textfont=sf]{subfig}
\usepackage{textcomp}
\usepackage{stfloats}
\usepackage{url}
\usepackage{verbatim}
\usepackage{graphicx}
\usepackage{cite}
\usepackage{multirow}
\usepackage{amsmath}
\usepackage{amsfonts}
\usepackage{breqn}
\usepackage{hyperref}

\begin{document}

\title{How Deep Learning Sees the World: A Survey on Adversarial Attacks \& Defenses}

\author{Joana C. Costa, Tiago Roxo, Hugo Proença,~\IEEEmembership{Senior Member,~IEEE,} Pedro R. M. Inácio,~\IEEEmembership{Senior Member,~IEEE}
\thanks{The authors are with Instituto de Telecomunicações, Universidade da Beira Interior, Portugal.}
\thanks{Manuscript received XX, 2023; revised XX, 2023.}}

\markboth{arXiv,~may~2023}%
{Costa \MakeLowercase{\textit{et al.}}: How Does Deep Learning See the World? A Survey on Adversarial Attacks \& Defenses}


\maketitle

\begin{abstract}

Deep Learning is currently used to perform multiple tasks, such as object recognition, face recognition, and natural language processing. However, Deep Neural Networks (DNNs) are vulnerable to perturbations that alter the network prediction (adversarial examples), raising concerns regarding its usage in critical areas, such as self-driving vehicles, malware detection, and healthcare. This paper compiles the most recent adversarial attacks, grouped by the attacker capacity, and modern defenses clustered by protection strategies. We also present the new advances regarding Vision Transformers, summarize the datasets and metrics used in the context of adversarial settings, and compare the state-of-the-art results under different attacks, finishing with the identification of open issues.

\end{abstract}


\section{Introduction}
\label{sec:intro}

\IEEEPARstart{M}{achine} Learning (ML) algorithms have been able to solve various types of problems, namely highly complex ones, through the usage of Deep Neural Networks (DNNs)~\cite{goodfellow2016deep}, achieving results similar to, or better than, humans in multiple tasks, such as object recognition~\cite{liu2020object,zhang2019object}, face recognition~\cite{masi2018face,wang2021face}, and natural language processing~\cite{wolf2020nlp,otter2020nlp}. These networks have also been employed in critical areas, such as self-driving vehicles~\cite{Maqueda2018CVPR,Ndikumana2021SelfDrive}, malware detection~\cite{Zhenlong2014Malware,Vinayakumar2019Malware}, and healthcare~\cite{zhou2020healthcare,liang2014healthcare}, whose application and impaired functioning can severely impact their users.

Promising results shown by DNNs lead to the sense that these networks could correctly generalize in the local neighborhood of an input (image). These results motivate the adoption and integration of these networks in real-time image analysis, such as traffic sign recognition and vehicle segmentation, making malicious entities target these techniques. However, it was discovered that DNNs are susceptible to small perturbations in their input~\cite{Szegedy2014IntriguingPO}, which entirely alter their prediction, making it harder for them to be applied in critical areas. These perturbations have two main characteristics: 1) invisible to the Human eye or slight noise that does not alter Human prediction; and 2) significantly increase the confidence of erroneous output, the DNNs predict the wrong class with higher confidence than all other classes. As a result of these assertions, the effect of the perturbations has been analyzed with more focus on object recognition, which will also be the main target of this survey.

Papernot \textit{et al.}~\cite{papernot2016limitations} distinguishes four types of adversaries depending on the information they have access to: (i) training data and network architecture, (ii) only training data or only network, (iii) oracle, and (iv) only pairs of input and output. In almost all real scenarios, the attacker does not have access to the training data or the network architecture, which diminishes the strength of the attack performed on a network, leaving the adversary with access to the responses given by the network, either by asking questions directly to it or by having pairs of input and prediction. Furthermore, the queries to a model are usually limited or very expensive~\cite{google2022pricing}, making it harder for an attacker to produce adversarial examples.

Multiple mechanisms~\cite{Liao2018DefenseAA,Samangouei2018DefenseGANPC,Zhang2019DefenseAA,Huang2021Defense} were proposed to defend against legacy attacks, already displaying their weakened effect when adequately protected, which are clustered based on six different domains in this survey. Regardless of the attacks and defenses already proposed, there is no assurance about the effective robustness of these networks and if they can be trusted in critical areas, clearly raising the need to make the DNNs inherently robust or easy to be updated every time a new vulnerability is encountered. This motivates the presented work, whose main contributions are summarized as follows:

\begin{itemize}
    \item We present the most recent adversarial attacks grouped by the adversary capacity, accompanied by an illustration of the differences between black-box and white-box attacks;
    \item We propose six different domains for adversarial defense grouping, assisted by exemplificative figures of each of these domains, and describe the effects of adversarial examples in ViTs;
    \item We detail the most widely used metrics and datasets, present state-of-the-art results on CIFAR-10, CIFAR-100, and ImageNet, and propose directions for future works.
\end{itemize}

The remaining of the paper is organized as follows: Section~\ref{sec:background} provides background information; Section~\ref{sec:rel_surveys} compares this review with others; Section~\ref{sec:adv_att} presents the set of adversarial attacks; Section~\ref{sec:adv_def} shows a collection of defenses to overcome these attacks; Section~\ref{sec:sota_datasets} displays the commonly used datasets; Section~\ref{sec:sota_metrics_results} lists and elaborates on metrics and presents state-of-the-art results; and Section~\ref{sec:fut_dir} presents future directions, with the concluding remarks included in Section~\ref{sec:conc}.

\section{Background for Adversarial Attacks}
\label{sec:background}

\subsection{Neural Network Architectures}

\begin{figure}[!tb]
    \centering
    \includegraphics[width=0.45\textwidth]{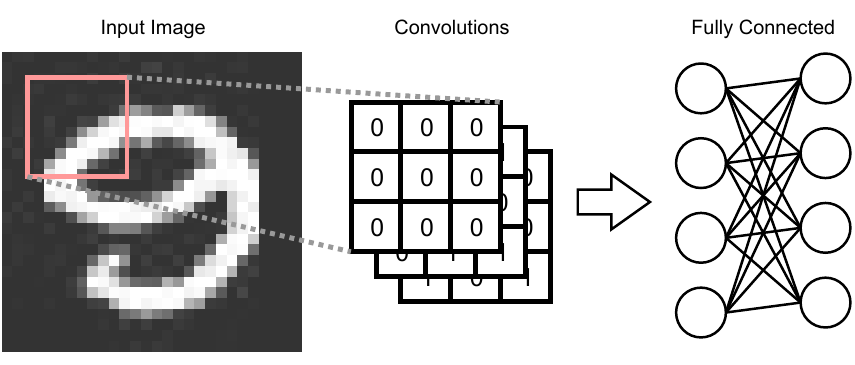}
    \caption{Schematic example of the Convolutional Neural Networks mechanism to classify images.}
    \label{fig:conv_nn_example}
\end{figure}

When an input image is fed into a CNN, it is converted into a matrix containing the numeric values representing the image or, if the image is colored, a set of matrices containing the numeric values for each color channel. Then, the Convolutions apply filters to these matrixes and calculate a set of reduced-size features. Finally, these features have an array format fed into the Fully Connected that classifies the provided image. Figure~\ref{fig:conv_nn_example} shows an elementary example of CNNs used to classify images.

\begin{figure}[!tb]
    \centering
    \includegraphics[width=0.45\textwidth]{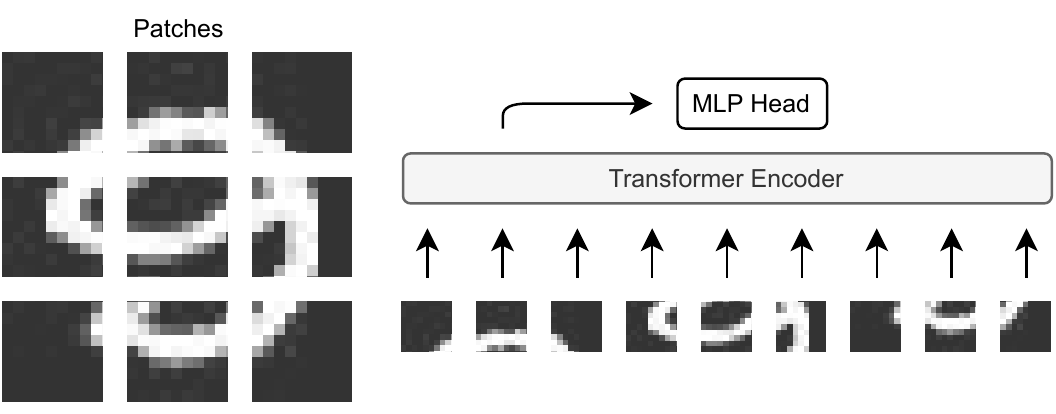}
    \caption{Schematic example of a simplified vision transformer used to classify images.}
    \label{fig:vision_transf_example}
\end{figure}

Contrary to the CNN, ViT does not receive the image as a whole as input; instead, it is pre-processed to be divided into Patches, which are smaller parts of the original image, as displayed in Figure~\ref{fig:vision_transf_example}. These Patches are not fed randomly to the Transformer Encoder, they are ordered by their position, and both the Patches and their position are fed into the Transformer Encoder. Finally, the output resulting from the Transformer Encoder is fed into the Multi-Layer Perceptron (MLP) Head that classifies the image.

\subsection{Adversarial Example}

Misclassification might be justified if the object contained in the image is not visible even to Humans. However, adversarial examples do not fit this scope. These examples add a perturbation to an image that causes the DNNs to misclassify the object in the image, yet Humans can correctly classify the same object.

The adversarial attacks described throughout this survey focus on identifying the adversarial examples that make DNNs misclassify. These attacks identify specific perturbations that modify the DNN classification while being correctly classified by Humans. The calculation of these perturbations is an optimization problem formally defined as:

\begin{equation}
\arg \min_{\delta \textbf{X}} \| \delta_{\textbf{X}} \| \textbf{ s.t. } \textbf{f}(\textbf{X} + \delta_{\textbf{X}}) = \textbf{Y}\ast,
\end{equation}
where $f$ is the is the classifier, $\delta_{\textbf{X}}$ is the perturbation, $\textbf{X}$ is the original/benign image, and $\textbf{Y}\ast$ is the adversarial output. Furthermore, the adversarial example is defined as:

\begin{equation}
\textbf{X}\ast = \textbf{X} + \delta_{\textbf{X}},
\end{equation}
where $\textbf{X}\ast$ is the adversarial image. 

\begin{figure}[!tb]
    \centering
    \includegraphics[width=0.4\textwidth]{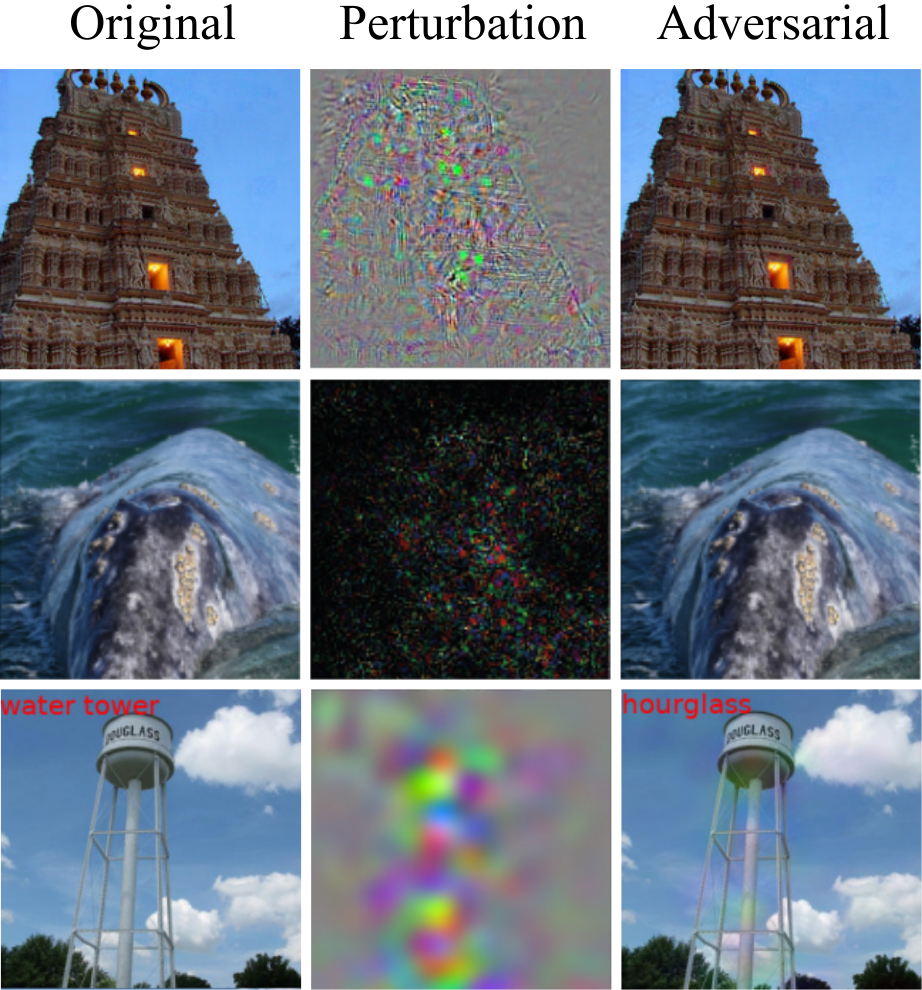}
    \caption{Adversarial Examples created using different state-of-the-art adversarial attacks. The first column represents the original image; the second represents the perturbation used to generate the adversarial images in the third column. The images were resized for better visualization. Images withdrawn from~\cite{Szegedy2014IntriguingPO,MoosaviDezfooli2016DeepFoolAS,Dabouei2020SmoothFoolAE}. The first perturbation follows the edges of the building, the second is concentrated in the area of the whale, and the third is more smooth and greater in area.}
    \label{fig:adversarial_examples}
\end{figure}

Figure~\ref{fig:adversarial_examples} displays adversarial examples generated using different attacks. Mainly, the first row is the Limited-memory Broyden-Fletcher-Goldfarb-Shanno (L-BFGS)~\cite{Szegedy2014IntriguingPO} attack, the second row is the DeepFool~\cite{MoosaviDezfooli2016DeepFoolAS} attack, and the third row is the SmoothFool~\cite{Dabouei2020SmoothFoolAE} attack. When observing the L-BFGS, the perturbation applies noise to almost the entirety of the adversarial image. The DeepFool attack only perturbs the area of the whale but not all the pixels in that area. Finally, the SmoothFool attack slightly disturbs the pixels in the area of the image. These three attacks display the evolution of adversarial attacks in decreasing order of detectability and, consequently, increasing order of strength.

To limit the noise that each perturbation can add to an image, the adversarial attacks are divided into $L_0$, $L_2$, and $L_p$ norms, known as \textit{Vector Norms}. Furthermore, commonly used terminologies in the context of adversarial examples are defined in Table~\ref{tab:terminologies}.

\subsection{Vector Norms and $\epsilon$ Constraint}
\label{sec:vec-norm-eps-cons}

Vector Norms are functions that take a vector as input and output a positive value (scalar). These functions are essential to ML and allow the backpropagation algorithms to compute the loss value as a scalar. The family of these functions is known as the \textit{p-norm}, and, in the context of adversarial attacks, the considered values for $p$ are $0$, $2$, and $\infty$.

\textbf{$L_0$ norm} consists of counting the number of non-zero elements in the vector and is formally given as:
\begin{equation}
||x||_0 = (|x_1|^0 + |x_2|^0 + ... + |x_n|^0),
\end{equation}
where $x_1$ to $x_n$ are the elements of the vector $x$.

\textbf{$L_2$ norm}, also known as the Euclidean distance, measures the vector distance to the origin and is formally defined as:
\begin{equation}
||x||_2 = (|x_1|^2 + |x_2|^2 + ... + |x_n|^2)^{\frac{1}{2}},
\end{equation}
where $x_1$ to $x_n$ are the elements of the vector $x$.

\textbf{$L_\infty$ norm} represents the maximum hypothetical value that $p$ can have and returns the absolute value of the element with the largest magnitude, formally as:
\begin{equation}
||x||_\infty = \max_{i}|x_i|,
\end{equation}
where $x_i$ is each element of the vector $x$.

\begin{figure}[!tb]
    \centering
    \includegraphics[width=0.48\textwidth]{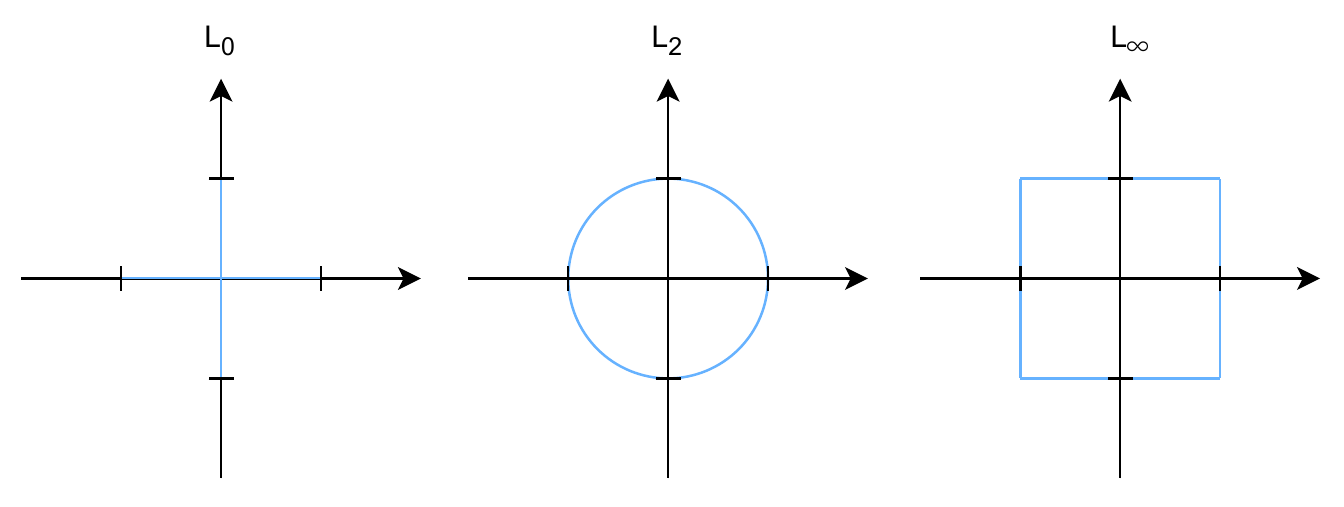}
    \caption{Geometric representation of the $l_0$, $l_2$, and $l_\infty$ norms, from left to right, respectively.}
    \label{fig:geometric_norms}
\end{figure}

A geometric representation of the area of exploitation for the three considered p-norm is displayed in Figure~\ref{fig:geometric_norms}. One relevant property of the p-norm is: the higher $p$ is, the more important the contribution of large errors; the lower $p$ is, the higher the contribution of small errors. This translates into a large $p$ benefiting small maximal errors (minimal perturbations along multiple pixels) and a small $p$ encouraging larger spikes in fewer places (abrupt perturbations along minimal pixels). Therefore, $l_2$ and $l_0$ attacks have greater detectability than $l_\infty$ attacks, with the latter being more threatening.

Another constraint normally seen in the context of adversarial attacks is $\epsilon$, which is a constant that controls the amount of noise, via generated perturbation, that can be added to an image. Usually, it is a tiny number and varies depending on the used dataset, decreasing when the task increases in difficulty. According to the literature, for MNIST, $\epsilon=0.1$, for CIFAR-10 and CIFAR-100, $\epsilon=8/255$, and for ImageNet, $\epsilon=4/255$.

\subsection{Adversary Goals and Capacity}

Besides the restriction imposed by the different Vector Norms, the adversarial attacks are also divided by their impact on the networks. Depending on the goals of the attacker, the designation is as follows:
\begin{itemize}
    \item \textbf{Confidence Reduction}, the classifier outputs the original label with less confidence; 
    \item \textbf{Untargeted}, the classifier outputs any class besides the original label;
    \item \textbf{Targeted}, the classifier outputs a particular class besides the original label.
\end{itemize}

Another important aspect of adversarial attacks is the amount of knowledge the attacker has access to. As defined by Papernot \textit{et al.}~\cite{papernot2016limitations}, who proposed the first threat model for deep learning, the attackers can have access to: 1) data training and network architecture; 2) only network architecture; 3) only data training; 4) an oracle that replies to all the inputs given; and 5) only have pairs of input and corresponding output (samples). However, to simplify this classification, these capacities were divided into:
\begin{itemize}
    \item \textbf{White-box}, which considers that the attacker has access to either the architecture or data;
    \item \textbf{Black-box}, when the attacker can only access samples from an oracle or in pairs of input and output.
\end{itemize}

The attackers goals and capacity are essential to classify the strength of an attack. For example, the easiest is a Confidence Reduction White-box attack, and the strongest is a Targeted Black-box attack.

\begin{table}[!tb]
    \centering
    \caption{Common terminologies used in the context of adversarial attacks and their definition.}
    \begin{tabular}{@{}m{1.8cm}|m{5.5cm}}
        \textbf{Terminology} & \textbf{Definition} \\
        \hline\hline
        Original/ Clean Example & Original image presented in a dataset \\
        \hline
        Adversarial/ Perturbed Example & Image that an adversary has manipulated to fool the classifier \\
        \hline
        Perturbation & Set of changes (for each pixel and color channel) that are performed on the image \\
        \hline
        Adversarial Attack & Technique used to calculate the perturbation that generates an adversarial example \\
        \hline
        Transferability & Capability of an adversarial example being transferred from a known network to an unknown network \\
        \hline
        White-box & Attacks that have access to DNN weights and datasets \\
        \hline
        Black-box & Attacks that do not have access to the DNN weights and datasets\\
        \hline
        Adversarial Training & Inclusion of adversarial examples in the training phase of the model \\
        \hline
    \end{tabular}
    \label{tab:terminologies}
\end{table}

\section{Related Surveys}
\label{sec:rel_surveys}

The first attempt to summarize and display the recent developments in this area was made by Akhtar and Mian~\cite{Akhtar2018Threat}. These authors studied adversarial attacks in computer vision, extensively referring to attacks for classification and providing a brief overview of attacks beyond the classification problem. Furthermore, the survey presents a set of attacks performed in the real world and provides insight into the existence of adversarial examples. Finally, the authors present the defenses distributed through three categories: modified training or input, modifying networks, and add-on networks.

From a broader perspective, Liu \textit{et al.}~\cite{Liu2018Threats} studied the security threats and possible defenses in ML scope, considering the different phases of an ML algorithm. For example, the training phase is only susceptible to poisoning attacks; however, the testing phase is vulnerable to evasion, impersonation, and inversion attacks, making it harder to defend. The authors provide their insight on the currently used techniques. Additionally, focusing more on the object recognition task, Serban \textit{et al.}~\cite{Serban2020Adversarial} extensively analyzed the adversarial attacks and defenses proposed under this context, providing conjectures for the existence of adversarial examples and evaluating the capacity of adversarial examples transferring between different DNNs.

Qui \textit{et al.}~\cite{qiu2019review} extensively explains background concepts in Adversarial Attacks, mentioning adversary goals, capabilities, and characteristics. It also displays applications for adversarial attacks and presents some of the most relevant adversarial defenses. Furthermore, it explains a set of attacks divided by the stage in which they occur, referring to the most relevant attacks.

Xu \textit{et al.}~\cite{xu2020adversarial} also describes background concepts, describing the adversary goals and knowledge. This review summarizes the most relevant adversarial attacks at the time of that work and presents physical world examples. Furthermore, the authors present a batch of defenses grouped by the underlying methodology. Finally, there is an outline of adversarial attacks in graphs, text, and audio networks, culminating in the possible applications of these attacks.

Chakraborty \textit{et al.}~\cite{chakraborty2021survey} provides insight into commonly used ML algorithms and presents the adversary capabilities and goals. The presented adversarial attacks are divided based on the stage of the attack (train or test). Additionally, the authors present relevant defenses used in adversarial settings.

\begin{table*}[!tb]
    \centering
    \caption{Characteristics shown in state-of-the-art surveys on Adversarial Attacks.}
    \begin{tabular}{@{}ccccccccccc}
        \multirow{2}{*}{Survey} & \multirow{2}{*}{Year} 
        & White \& & Survey & Grouping of & Future & Datasets & Metrics and & State-of-the-art & Vision \\
        &  
        & Black-Box & Comparison & Defenses & Directions & Overview & Architectures & Comparison & Transformers \\
        \hline\hline
        Akhtar and Mian~\cite{Akhtar2018Threat} & 2018 & \checkmark & $\times$ & \checkmark & \checkmark & $\times$ & $\times$ & $\times$ & $\times$ \\

        Qiu \textit{et al.}~\cite{qiu2019review} & 2019 & \checkmark & $\times$ & $\times$  & $\times$ & $\times$ & $\times$  & $\times$ & $\times$ \\
        
        Serban \textit{et al.}~\cite{Serban2020Adversarial} & 2020 & \checkmark & $\times$ & \checkmark  & \checkmark & $\times$ & $\times$  & $\times$ & $\times$ \\

        Xu \textit{et al.}~\cite{xu2020adversarial} & 2020 & \checkmark & $\times$ & \checkmark  & $\times$ & $\times$ & $\times$  & $\times$ & $\times$ \\
        
        Chakraborty \textit{et al.}~\cite{chakraborty2021survey} & 2021 & \checkmark & $\times$ & $\times$  & $\times$ & $\times$ & $\times$  & $\times$ & $\times$ \\

        Long \textit{et al.}~\cite{long2022survey} & 2022 & \checkmark & \checkmark & $\times$  & \checkmark & $\times$ & $\times$  & $\times$ & $\times$ \\

        Liang \textit{et al.}~\cite{liang2022adversarial} & 2022 & \checkmark & $\times$ & \checkmark  & \checkmark & $\times$ & $\times$ & $\times$ & $\times$ \\

        Zhou \textit{et al.}~\cite{zhou2022adversarial} & 2022 & \checkmark & $\times$ & \checkmark  & \checkmark & \checkmark & $\times$ & $\times$ & $\times$ \\

        \hline
        This survey & 2023 & \checkmark & \checkmark & \checkmark & \checkmark & \checkmark & \checkmark & \checkmark & \checkmark \\
        
    \end{tabular}
    \label{tab:survey_comparison}
\end{table*}

Long \textit{et al.}~\cite{long2022survey} discusses a set of preliminary concepts of Computer Vision and adversarial context, providing a set of adversarial attacks grouped by adversary goals and capabilities. Finally, the authors provide a set of research directions that readers can use to continue the development of robust networks.

Liang \textit{et al.}~\cite{liang2022adversarial} discuss the most significant attacks and defenses in the literature, with the latter being grouped by the underlying technique. This review finishes with a presentation of the challenges currently existing in the adversarial context.

More recently, Zhou \textit{et al.}~\cite{zhou2022adversarial} provides insight into Deep Learning and Threat Models, focusing on the Cybersecurity perspective. Therefore, the authors identify multiple stages based on Advanced Persistent Threats and explain which adversarial attacks are adequate for each stage. Similarly, the same structure is followed to present the appropriate defenses for each stage. Furthermore, this survey presents the commonly used datasets in adversarial settings and provides a set of future directions from a Cybersecurity perspective.

From the analysis of the previous surveys, some concepts have already been standardized, such as adversary goals and capabilities and the existence of adversarial attacks and defenses. However, due to the recent inception of this area, there still needs to be more standardization in datasets and metrics. Therefore, with this survey, we also analyze datasets and metrics to provide insight to novice researchers. Furthermore, this survey consolidates the state-of-the-art results and identifies which datasets can be further explored. Finally, similarly to other reviews, this paper provides a set of future directions that researchers and practitioners can follow to start their work. A comparison between the several surveys discussed in this section is summarized in Table~\ref{tab:survey_comparison}.

\section{Adversarial Attacks}
\label{sec:adv_att}

\begin{figure*}[!tb]
    \centering
    \includegraphics[width=.96\textwidth]{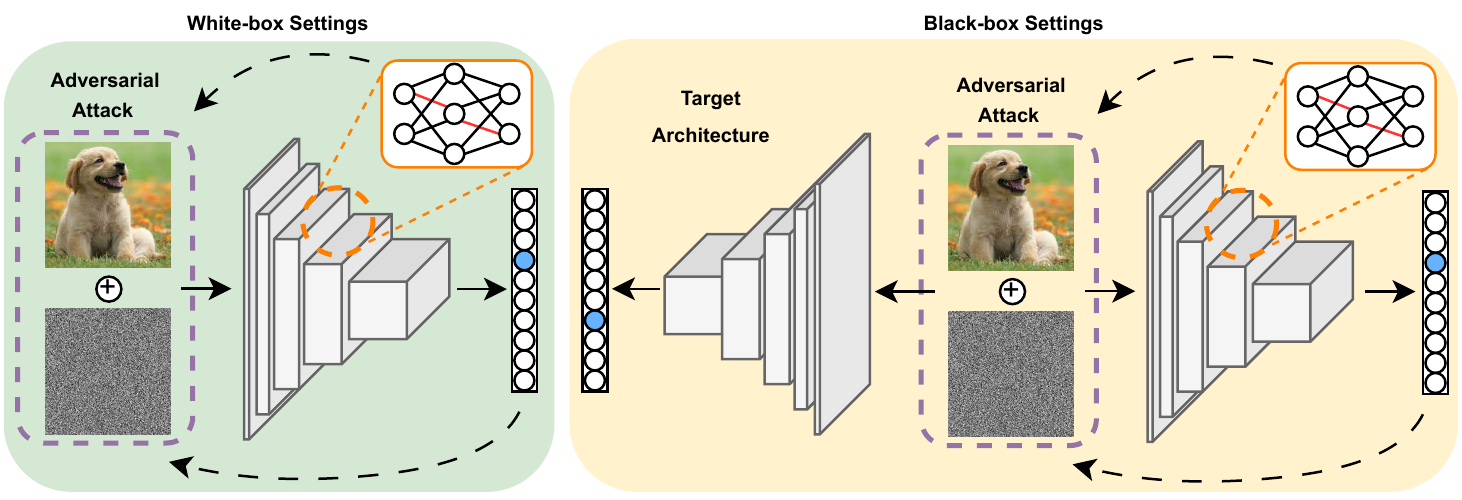}
    \caption{Schematic overview of an Adversarial Attack under White-box Settings (left) and Black-box Settings (right). The first one uses the classifier predictions and network gradients to create perturbations (similar to noise), which can fool this classifier. These perturbations are added to the original images, creating adversarial images, which are fed to the network and cause misclassification. In the Black-box Settings, the same process is applied to a known classifier, and the obtained images are used to attack another classifier (represented as Target Architecture).}
    \label{fig:white_vs_black_box}
\end{figure*}

Adversarial attacks are commonly divided by the amount of knowledge the adversaries have access to, white-box and black-box, as can be seen in Figure~\ref{fig:white_vs_black_box}.

\subsection{White-box Settings}

Adversarial examples were first proposed by Szegedy \textit{et al.}~\cite{Szegedy2014IntriguingPO}, which discovered that DNNs do not generalize well in the vicinity of an input. The same authors proposed \textbf{L-BFGS}, the first adversarial attack, to create adversarial examples and raised awareness in the scientific community for this generalization problem.

\textbf{Fast Gradient Sign Method} (FGSM)~\cite{Goodfellow2015ExplainingAH} is a one-step method to find adversarial examples, which is based on the linear explanation for the existence of adversarial examples, and is calculated using the model cost function, the gradient, and the radius epsilon. This attack is formally defined as:
\begin{equation}
    x - \epsilon \cdot \text{sign}(\nabla\text{loss}_{F,t}(x)),
\end{equation}
where $x$ is the original image, $\epsilon$ is the amount of changes to the image, and $t$ is the target label. The value for $\epsilon$ should be very small to make the attack undetectable.

\textbf{Jacobian-based Saliency Maps} (JSM)~\cite{papernot2016limitations} explore the forward derivates to calculate the model gradients, replacing the gradient descent approaches, and discover which input regions are likely to yield adversarial examples. Then it uses saliency maps to construct the adversarial saliency maps, which display the features the adversary must perturb. Finally, to prove the effectiveness of JSM, only the adversarial examples correctly classified by humans were used to fool neural networks.

\textbf{DeepFool}~\cite{MoosaviDezfooli2016DeepFoolAS} is an iterative attack that stops when the minimal perturbation that alters the model output is found, exploiting its decision boundaries. It finds the minimum perturbation for an input $x_0$, corresponding to the vector orthogonal to the hyperplane representing the decision boundary.

Kurakin \textit{et al.}~\cite{Kurakin2017AdversarialEI} was the first to demonstrate that adversarial examples can also exist in the physical world, by using three different methods to generate the adversarial examples. \textbf{Basic Iterative Method} (BIM) applies the FGSM multiple times with a small step size between iterations and clips the intermediate values after each step. \textbf{Iterative Least-likely Class Method} (ILCM) uses the least-likely class, according to the prediction of the model, as the target class and uses BIM to calculate the adversarial example that outputs the target class.

\textbf{Carlini and Wagner} (C\&W)~\cite{carlini2017towards} attack is one of the most powerful attacks, which uses three different vector norms: 1) the $L_2$ attack uses a smoothing of clipped gradient descent approach, displaying low distortion; 2) the $L_0$ attack uses an iterative algorithm that, at each iteration, fixes the pixels that do not have much effect on the classifier and finds the minimum amount of pixels that need to be altered; and 3) the $L_\infty$ attack also uses an iterative algorithm with an associated penalty, penalizing every perturbation that exceeds a predefined value, formally defined as:
\begin{equation}
        \text{min} \quad c \cdot f(x + \delta) + \sum_i [(\delta_i - \tau)^+],
\end{equation}
where $\delta$ is the perturbation, $\tau$ is the penalty threshold (initially 1, decreasing in each iteration), and $c$ is a constant. The value for $c$ starts as a very low value (\textit{e.g.}, $10^{-4}$), and each time the attack fails, the value for $c$ is doubled. If $c$ exceeds a threshold (\textit{e.g.}, $10^{10}$), it aborts the search.

\textbf{Gradient Aligned Adversarial Subspace} (GAAS)~\cite{Tramr2017TheSO} is an attack that directly estimates the dimensionality of the adversarial subspace using the first-order approximation of the loss function. Through the experiments, GAAS proved the most successful at finding many orthogonal attack directions, indicating that neural networks generalize linearly.

\textbf{Projected Gradient Descent} (PGD)~\cite{Madry2018TowardsDL} is an iterative attack that uses saddle point formulation, viewed as an inner maximization problem and an outer minimization problem, to find a strong perturbation. It uses the inner maximization problem to find an adversarial version of a given input that achieves a high loss and the outer minimization problem to find model parameters that minimize the loss in the inner maximization problem. The saddle point problem used by PGD is defined as:
\begin{equation}
    \min_\theta \rho(\theta), \text{where}~\rho(\theta) = \mathbb{E}_{(x,y)\sim \mathcal{D}} \Bigl[ \max_{\delta \in \mathcal{S}} L(\theta, x + \delta, y) \Bigr],
\end{equation}
where $x$ is the original image, $y$ is the corresponding label, and $\mathcal{S}$ is the set of allowed perturbations.

\textbf{AdvGAN}~\cite{Xiao2018GeneratingAE} uses Generative Adversarial Networks (GAN)~\cite{Goodfellow2014GenerativeAN} to create adversarial examples that are realistic and have high attack success rate. The generator receives the original instance and creates a perturbation, the discriminator distinguishes the original instance from the perturbed instance, and the target neural network is used to measure the distance between the prediction and the target class.

Motivated by the inability to achieve a high success rate in black-box settings, the \textbf{Momentum Iterative FGSM} (MI-FGSM)~\cite{Dong2018BoostingAA} was proposed. It introduces momentum, a technique for accelerating gradient descent algorithms, into the already proposed Iterative FGSM (I-FGSM), showing that the attack success rate in black-box settings increases almost double that of previous attacks.

Croce and Hein~\cite{Croce2019SparseAI} noted that the perturbations generated by $l_0$ attacks are sparse and by $l_\infty$ attacks are smooth on all pixels, proposing \textbf{Sparse and Imperceivable Adversarial Attacks} (SIAA). This attack creates sporadic and imperceptible perturbations by applying the standard deviation of each color channel in both axis directions, calculated using the two immediate neighboring pixels and the original pixel.

\textbf{SmoothFool} (SF)~\cite{Dabouei2020SmoothFoolAE} is a geometry-inspired framework for computing smooth adversarial perturbations, exploiting the decision boundaries of a model. It is an iterative algorithm that uses DeepFool to calculate the initial perturbation and smoothly rectifies the resulting perturbation until the adversarial example fools the classifier. This attack provides smoother perturbations which improve the transferability of the adversarial examples, and their impact varies with the different categories in a dataset.

In the context of exploring the adversarial examples in the physical world, the \textbf{Adversarial Camouflage} (AdvCam)~\cite{Duan2020AdversarialCH}, which crafts physical-world adversarial examples that are legitimate to human observers, was proposed. It uses the target image, region, and style to perform a physical adaptation (creating a realistic adversarial example), which is provided into a target neural network to evaluate the success rate of the adversarial example.

\textbf{Feature Importance-aware Attack} (FIA)~\cite{Wang2021FeatureIT} considers the object-aware features that dominate the model decisions, using the aggregate gradient (gradients average concerning the feature maps). This approach avoids local optimum, represents transferable feature importance, and uses the aggregate gradient to assign weights identifying the essential features. Furthermore, FIA generates highly transferable adversarial examples when extracting the feature importance from multiple classification models.

\textbf{Meta Gradient Adversarial Attack} (MGAA)~\cite{Yuan2021MetaGA} is a novel architecture that can be integrated into any existing gradient-based attack method to improve cross-model transferability. This approach consists of multiple iterations, and, in each iteration, various models are samples from a model zoo to generate adversarial perturbations using the selected model, which are added to the previously generated perturbations. In addition, using multiple models simulates both white- and black-box settings, making the attacks more successful.

\subsection{Universal Adversarial Perturbations}

Moosavi-Dezfooli \textit{et al.}~\cite{MoosaviDezfooli2017UniversalAP} discovered that some perturbations are image-agnostic (universal) and cause misclassification with high probability, labeled as \textbf{Universal Adversarial Perturbations} (UAPs). The authors found that these perturbations also generalize well across multiple neural networks, by searching for a vector of perturbations that cause misclassification in almost all the data drawn from a distribution of images. The optimization problem that Moosavi-Dezfooli \textit{et al.} are trying to solve is the following:
\begin{equation}
    \Delta v_i \xleftarrow{} \textrm{arg} \min_r \| r \|_2 \quad \textrm{s.t.} \quad \hat{k}(x_i + v + r) \neq \hat{k}(x_i),
\end{equation}
where $\Delta v_i$ is the minimal perturbation to fool the classifier, $v$ is the universal perturbation, and $x_i$ is the original image. This optimization problem is calculated for each image in a dataset, and the vector containing the universal perturbation is updated.

The \textbf{Universal Adversarial Networks} (UAN)~\cite{Hayes2018LearningUA} are Generative Networks that are capable of fooling a classifier when their output is added to an image. These networks were inspired by the discovery of UAPs, which were used as the training set and can create perturbations for any given input, demonstrating more outstanding results than the original UAPs.

\subsection{Black-box Settings}

Specifically considering black-box setup, Ilyas \textit{et al.}~\cite{Ilyas2018BlackboxAA} define three realistic threat models that are more faithful to real-world settings: query-limited, partial information, and label-only settings. The first one suggests the development of query-efficient algorithms, using Natural Evolutionary Strategies to estimate the gradients used to perform the PGD attack. When only having the probabilities for the top-k labels, the algorithm alternates between blending in the original image and maximizing the likelihood of the target class and, when the attacker only obtains the top-k predicted labels, the attack uses noise robustness to mount a targeted attack.

\textbf{Feature-Guided Black-Box} (FGBB)~\cite{Wicker2018FeatureGuidedBS} uses the features extracted from images to guide the creation of adversarial perturbations, by using Scale Invariant Feature Transform. High probability is assigned to pixels that impact the composition of an image in the Human visual system and the creation of adversarial examples is viewed as a two-player game, where the first player minimizes the distance to an adversarial example, and the second one can have different roles, leading to minimal adversarial examples.

\textbf{Square Attack}~\cite{andriushchenko2020square} is an adversarial attack that does not need local gradient information, meaning that gradient masking does not affect it. Furthermore, this attack uses a randomized search scheme that selects localized square-shaped updates in random positions, causing the perturbation to be situated at the decision boundaries.

\subsection{Auto-Attack}

\textbf{Auto-Attack}~\cite{croce2020reliable} was proposed to test adversarial robustness in a parameter-free, computationally affordable, and user-independent way. As such, Croce \textit{et al.} proposed two variations of PGD to overcome suboptimal step sizes of the objective function, namely APGD-CE and APGD-DLR, for a step size-free version of PGD using cross-entropy (CE) and Difference of Logits Ratio (DLR) loss, respectively. DLR is a loss proposed by Croce \textit{et al.} which is both shift and rescaling invariant and thus has the same degrees of freedom as the decision of the classifier, not suffering from the issues of the cross-entropy loss~\cite{croce2020reliable}. Then, they combine these new PGD variations with two other existing attacks to create Auto-Attack, which is composed by:
\begin{itemize}
    \item APGD-CE, step size-free version of PGD on the cross-entropy;
    \item APGD-DLR, step size-free version of PGD on the DLR loss;
    \item Fast Adaptive Boundary (FAB)~\cite{croce2020minimally}, which minimizes the norm of the adversarial perturbations;
    \item Square~\cite{andriushchenko2020square} Attack, a query-efficient black-box attack.
\end{itemize}

Given the main motivation of the Auto-Attack proposal, the FAB attack is the targeted version of FAB~\cite{croce2020minimally} since the untargeted version computes each iteration of the Jacobian matrix of the classifier, which scales linearly with the number of classes of the dataset. Although this is feasible for datasets with a low number of classes (\textit{e.g.}, MNIST and CIFAR-10), it becomes both computationally and memory-wise challenging with an increased number of classes (\textit{e.g.}, CIFAR-100 and ImageNet).

As such, Auto-Attack is an ensemble of attacks with important fundamental properties: APGD is a white-box attack aiming at any adversarial example within an $L_{p}$-ball (Section~\ref{sec:vec-norm-eps-cons}), FAB minimizes the norm of the perturbation necessary to achieve a misclassification, and Square Attack is a score-based black-box attack for norm bounded perturbations which use random search and do not exploit any gradient approximation, competitive with white-box attacks~\cite{andriushchenko2020square}.

\section{Adversarial Defenses}
\label{sec:adv_def}

\subsection{Adversarial Training}

\begin{figure}[!tb]
    \centering
    \includegraphics[width=.48\textwidth]{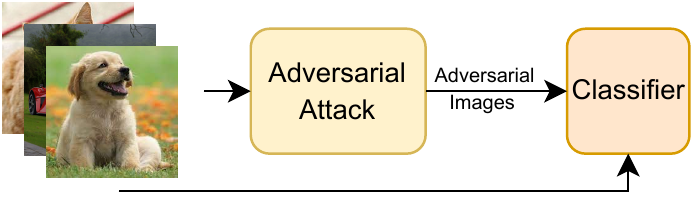}
    \caption{Schematic overview of Adversarial Training. A subset of the original images of a dataset is fed into an adversarial attack (\textit{e.g.}, PGD, FGSM, or C\&W), which creates adversarial images. Each batch contains original and adversarial images, with the Classifier being normally trained.}
    \label{fig:adv_train}
\end{figure}

Szegedy \textit{et al.}~\cite{Szegedy2014IntriguingPO} proposed that training on a mixture of adversarial and clean examples could regularize a neural network, as shown in Figure~\ref{fig:adv_train}. Goodfellow \textit{et al.}~\cite{Goodfellow2015ExplainingAH} evaluated the impact of \textbf{Adversarial Training} as a regularizer by including it in the objective function, showing that this approach is a reliable defense that can be applied to every neural network.

Kurakin \textit{et al.}~\cite{Kurakin2017AdversarialML} demonstrates that it is possible to perform adversarial training in more massive datasets (ImageNet), displaying that the robustness significantly increases for one-step methods. When training the model with one-step attacks using the ground-truth labels, the model has significantly higher accuracy on the adversarial images than on the clean images, an effect denominated as \textbf{Label Leaking}, suggesting that the adversarial training should not make use of the ground-truth labels.

Adversarial Training in large datasets implies using fast single-step methods, which converge to a degenerate global minimum, meaning that models trained with this technique remain vulnerable to black-box attacks. Therefore, \textbf{Ensemble Adversarial Training}~\cite{Tramr2018EnsembleAT} uses adversarial examples crafted on other static pre-trained models to augment the training data, preventing the trained model from influencing the strength of the adversarial examples.

\textbf{Shared Adversarial Training}~\cite{Mummadi2018SharedAdvT} is an extension of adversarial training aiming to maximize robustness against universal perturbations. It splits the mini-batch of images used in training into a set of stacks and obtains the loss gradients concerning these stacks. Afterward, the gradients for each stack are processed to create a shared perturbation that is applied to the whole stack. After every iteration, these perturbations are added and clipped to constrain them into a predefined magnitude. Finally, these perturbations are added to the images and used for adversarial training.

\textbf{TRadeoff-inspired Adversarial DEfense via Surrogate-loss minimization} (TRADES)~\cite{zhang2019theoretically} is inspired by the presumption that robustness can be at odds with accuracy~\cite{tsipras2018robustness, su2018robustness}. The authors show that the robust error can be tightly bounded by using natural error measured by the surrogate loss function and the likelihood of input features being close to the decision boundary (boundary error). These assumptions make the model weights biased toward natural or boundary errors.

Based on the idea that gradient magnitude is directly linked to model robustness, \textbf{Bilateral Adversarial Training} (BAT)~\cite{wang2019bilateral} proposes to perturb not only the images but also the manipulation of labels (adversarial labels) during the training phase. The adversarial labels are derived from a closed-form heuristic solution, and the adversarial images are generated from a one-step targeted attack.

Despite the popularity of adversarial training to defend models, it has a high cost of generating strong adversarial examples, namely for large datasets such as ImageNet. Therefore, \textbf{Free Adversarial Training} (Free-AT)~\cite{shafahi2019adversarial} uses the gradient information when updating model parameters to generate the adversarial examples, eliminating the previously mentioned overhead. 

Considering the same issue presented in Free-AT, the authors analyze Pontryagin's Maximum Principle~\cite{kopp1962pontryagin} of this problem and observe that the adversary update is only related to the first layer of the network. Thus, \textbf{You Only Propagate Once} (YOPO)~\cite{zhang2019you} only considers the first layer of the network for forward and backpropagation, effectively reducing the amount of propagation to one in each update.

\textbf{Misclassification Aware adveRsarial Training} (MART)~\cite{wang2020improving} is an algorithm that explicitly differentiates the misclassified and correctly classified examples during training. This proposal is motivated by the finding that different maximization techniques are negligible, but minimization ones are crucial when looking at the misclassified examples. 

\textbf{Defense against Occlusion Attacks} (DOA)~\cite{wu2019defending} is a defense mechanism that uses abstract adversarial attacks, Rectangular Occlusion Attack (ROA)~\cite{wu2019defending}, and applies the standard adversarial training. This attack considers including physically realizable attacks that are ``normal'' in the real world, such as eyeglasses and stickers on stop signs.

The proposal of \textbf{Smooth Adversarial Training} (SAT)~\cite{sitawarin2020improving} considers the evolution normally seen in curriculum learning, where the difficulty increases with time (age), using two difficulty metrics. These metrics are based on the maximal Hessian eigenvalue (H-SAT) and the softmax Probability (P-SAT), which are used to stabilize the networks for large perturbations while having high clean accuracy. In the same context, \textbf{Friendly Adversarial Training} (Friend-AT)~\cite{zhang2020attacks} minimizes the loss considering the least adversarial data (friendly) among the adversarial data that is confidently misclassified. This method can be employed by early stopping PGD attacks when performing adversarial training.

Contrary to the idea of Free-AT~\cite{shafahi2019adversarial}, \textbf{Cheap Adversarial Training} (Cheap-AT)~\cite{wong2020fast} proposes the use of weaker and cheaper adversaries (FGSM) combined with random initialization to train robust networks effectively. This method can be further accelerated by applying techniques that efficiently train networks.

In a real-world context, the attacks are not limited by the imperceptibility constraint ($\epsilon$ value); there are, in fact, multiple perturbations (for models) that have visible sizes. The main idea of \textbf{Oracle-Aligned Adversarial Training} (OA-AT)~\cite{addepalli2021towards} is to create a model that is robust to high perturbation bounds by aligning the network predictions with ones of an Oracle during adversarial training. The key aspect of OA-AT is the use of Learned Perceptual Image Patch Similarity~\cite{zhang2018unreasonable} to generate Oracle-Invariant attacks and convex combination of clean and adversarial predictions as targets for Oracle-Sensitive samples.

\textbf{Geometry-aware Instance-reweighted Adversarial Training} (GI-AT)~\cite{zhang2020geometry} has two foundations: 1) over-parameterized models still lack capacity; and 2) a natural data point closer to the class boundary is less robust, translating into assigning the corresponding adversarial data a larger weight. Therefore, this defense proposes using standard adversarial training, considering that weights are based on how difficult it is to attack a natural data point. 

Adversarial training leads to unfounded increases in the margin along decision boundaries, reducing clean accuracy. To tackle this issue, \textbf{Helper-based Adversarial Training} (HAT)~\cite{rade2021helper} incorporates additional wrongly labeled examples during training, achieving a good trade-off between accuracy and robustness.

As a result of the good results achieved by applying random initialization, \textbf{Fast Adversarial Training} (FAT)~\cite{chen2022efficient} performs randomized smoothing to optimize the inner maximization problem efficiently, and proposes a new initialization strategy, named backward smoothing. This strategy helps to improve the stability and robustness of a model using single-step robust training methods, solving the overfitting issue.

\subsection{Modify the Training Process}
Gu and Rigazio~\cite{Gu2015TowardsDN} proposed using three preprocessing techniques to recover from the \textbf{adversarial noise}, namely, noise injection, autoencoder, and denoising autoencoder, discovering that the adversarial noise is mainly distributed in the high-frequency domain. Solving the adversarial problem corresponds to encountering adequate training techniques and objective functions to increase the distortion of the smallest adversarial examples.

\begin{figure}[!tb]
    \centering
    \includegraphics[width=.46\textwidth]{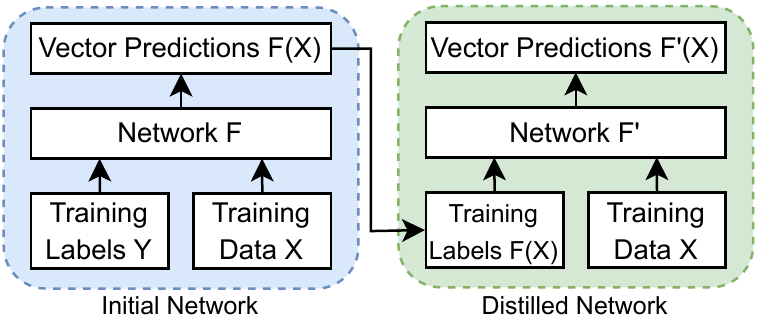}
    \caption{Method proposed by Defensive Distillation~\cite{Papernot2016DistillationAA}. An Initial Network is trained on the dataset images and labels (discrete values). Then, the predictions given by the Initial Network are fed into another network, replacing the dataset labels. These predictions are continuous values, making the Distilled Network more resilient to adversarial attacks.}
    \label{fig:defensive_distillation}
\end{figure}

Another defense against adversarial examples is \textbf{Defensive Distillation}~\cite{Papernot2016DistillationAA}, which uses the predictions from a previously trained neural network, as displayed in Figure~\ref{fig:defensive_distillation}. This approach trains the initial neural network with the original training data and labels, producing the probability of the predictions, which replace the original training labels to train a smaller and resilient distilled network. Additionally, to improve the results obtained by Defensive Distillation, Papernot and McDaniel~\cite{Papernot2017ExtendingDD} propose to change the vector used to train the distilled network by combining the original label with the first model uncertainty.

To solve the vulnerabilities of the neural network to adversarial examples, the \textbf{Visual Causal Feature Learning}~\cite{Chalupka2015VisualCF} method uses causal reasoning to perform data augmentation. This approach uses manipulator functions that return an image similar to the original one with the desired causal effect.

\textbf{Learning with a Strong Adversary}~\cite{Huang2015LearningWA} is a training procedure that formulates as a min-max problem, making the classifier inherently robust. This approach considers that the adversary applies perturbations to each data point to maximize the classification error, and the learning procedure attempts to minimize the misclassification error against the adversary. The greatest advantage of this procedure is the significant increase in robustness while maintaining clean high accuracy.

Zheng \textit{et al.}~\cite{Zheng2016ImprovingTR} proposes the use of compression, rescaling, and cropping in benign images to increase the stability of DNNs, denominated as \textbf{Image Processing}, without changing the objective functions. A Gaussian perturbation sampler perturbs the benign image, which is fed to the DNN, and its feature representation of benign images is used to 1) minimize the standard CE loss; and 2) minimize the stability loss.

Zantedeschi \textit{et al.}~\cite{Zantedeschi2017Efficient} explored the standard architectures, which usually employ Rectified Linear Units (ReLU)~\cite{Agarap2018ReLU,Hahnloser2000Digital} to ease the training process, and discovered that this function makes a small perturbation in the input accumulate with multiple layers (unbounded). Therefore, the authors propose the use of bounded ReLU (BReLU)~\cite{Liew2016BoundedReLU} to prevent this accumulation and \textbf{Gaussian Data Augmentation} to perform data augmentation.

Zhang and Wang~\cite{Zhang2019DefenseAA} suggest that adversarial examples are generated through \textbf{Feature Scattering} (FS) in the latent space to avoid the label leaking effect, which considers the inter-example relationships. The adversarial examples are generated by maximizing the feature-matching distance between the clean and perturbed examples, FS produces a perturbed empirical distribution, and the DNN performs standard adversarial training.

PGD attack causes the internal representation to shift closer to the ``false'' class, \textbf{Triplet Loss Adversarial} (TLA)~\cite{mao2019metric} includes an additional term in the loss function that pulls natural and adversarial images of a specific class closer and the remaining classes further apart. This method was tested with different samples: Random Negative (TLA-RN), which refers to a randomly sampled negative example, and Switch Anchor (TLA-SA), which sets the anchor as a natural example and the positive to be adversarial examples.

Kumari \textit{et al.}~\cite{kumari2019harnessing} analyzes the previously adversarial-trained models to test their vulnerability against adversarial attacks at the level of latent layers, concluding that the latent layer of these models is significantly vulnerable to adversarial perturbations of small magnitude. \textbf{Latent Adversarial Training} (LAT)~\cite{kumari2019harnessing} consists of finetuning adversarial-trained models to ensure robustness at the latent level.

\textbf{Curvature Regularization} (CR)~\cite{moosavi2019robustness} minimizes the curvature of the loss surface, which induces a more "natural" behavior of the network. The theoretical foundation behind this defense uses a locally quadratic approximation that demonstrates a strong relation between large robustness and small curvature. Furthermore, the proposed regularizer confirms the assumption that exhibiting quasi-linear behavior in the proximity of data points is essential to achieve robustness.

\textbf{Unsupervised Adversarial Training} (UAT)~\cite{alayrac2019labels} enables the training with unlabeled data considering two different approaches, UAT with Online Target (UAT-OT) that minimizes a differentiable surrogate of the smoothness loss, and UAT with Fixed Targets (UAT-FT) that trains an external classifier to predict the labels on the unsupervised data and uses its predictions as labels.

\textbf{Robust Self-Training} (RST)~\cite{carmon2019unlabeled}, an extension of Self-Training~\cite{scudder1965probability,cohen2019certified}, uses a standard supervised training to obtain pseudo-labels and then feeds them into a supervised training algorithm that targets adversarial robustness. This approach bridges the gap between standard and robust accuracy, using the unlabeled data, achieving high robustness using the same number of labels as required for high standard accuracy.

\textbf{SENSEI}~\cite{Gao2020FuzzTB} and \textbf{SENSEI-SA}~\cite{Gao2020FuzzTB} use the methodologies employed in software testing to perform data augmentation, enhancing the robustness of DNNs. SENSEI implements the strategy of replacing each data point with a suitable variant or leaving it unchanged. SENSEI-SA improves the previous one by identifying which opportunities are suitable for skipping the augmentation process.

\begin{figure}[!tb]
    \centering
    \includegraphics[width=.46\textwidth]{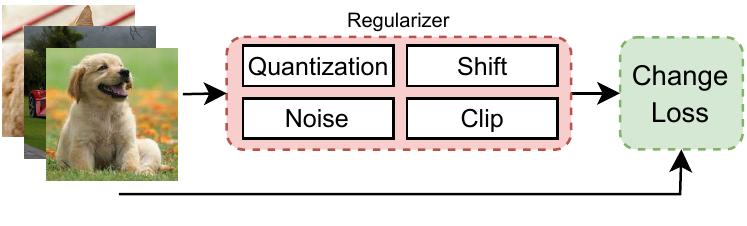}
    \caption{Schematic overview of the Bit Plane Feature Consistency~\cite{Addepali2020BPFC}. This method applies multiple operations to input images, simulating adversarial images. Then, the loss is changed to include a regularizer (new term), which compares the original images with these manipulated images.}
    \label{fig:bpfc}
\end{figure}

\textbf{Bit Plane Feature Consistency} (BPFC)~\cite{Addepali2020BPFC} regularizer forces the DNNs to give more importance to the higher bit planes, inspired by the Human visual system perception. This regularizer uses the original image and a preprocessed version to calculate the $l_2$ norm between them and regularize the loss function, as the scheme shown in Figure~\ref{fig:bpfc}.

\textbf{Adversarial Weight Perturbation} (AWP)~\cite{wu2020adversarial} explicitly regularizes the flatness of weight loss landscape and robustness gap, using a double-perturbation mechanism that disturbs both inputs and weights. This defense boosts the robustness of multiple existing adversarial training methods, confirming that it can be applied to other methods.

\textbf{Self-Adaptive Training} (SAT)~\cite{huang2020self} dynamically calibrates the training process with the model predictions without extra computational cost, improving the generalization of corrupted data. In contrast with the double-descent phenomenon, SAT exhibits a single-descent error-capacity curve, mitigating the overfitting effect.

\textbf{HYDRA}~\cite{sehwag2020hydra} is another technique that explores the effects of pruning on the robustness of models, which proposes using pruning techniques that are aware of the robust training objective, allowing this objective to guide the search for connections to prune. This approach reaches compressed models that are state-of-the-art in standard and robust accuracy.

Based on the promising results demonstrated by previous distillation methods, the \textbf{Robust Soft Label Adversarial Distillation} (RSLAD)~\cite{Zi2021RSLAD} method uses soft labels to train robust small student DNNs. This method uses the Robust Soft Labels (RSLs) produced by the teacher DNN to supervise the student training on natural and adversarial examples. An essential aspect of this method is that the student DNN does not access the original complex labels through the training process.

The most sensitive neurons in each layer make significant non-trivial contributions to the model predictions under adversarial settings, which means that increasing adversarial robustness stabilizes the most sensitive neurons. \textbf{Sensitive Neuron Stabilizing} (SNS)~\cite{Zhang2021InterpretingAI} includes an objective function dedicated explicitly to maximizing the similarities of sensitive neuron behaviors when providing clean and adversarial examples.

\textbf{Dynamic Network Rewiring} (DNR)~\cite{kundu2021dnr} generates pruned DNNs that have high robust and standard accuracy, which employs a unified constrained optimization formulation using a hybrid loss function that merges ultra-high model compression with robust adversarial training. Furthermore, the authors propose a one-shot training method that achieves high compression, standard accuracy, and robustness, which has a practical inference 10 times faster than traditional methods. 

\textbf{Manifold Regularization for Locally Stable} (MRLS)~\cite{jin2020manifold} DNNs exploit the continuous piece-wise linear nature of ReLU to learn a function that is smooth over both predictions and decision boundaries. This method is based on approximating the graph Laplacian when the data is sparse.

Inspired by the motivation behind distillation, \textbf{Learnable Boundary Guided Adversarial Training} (LBGAT)~\cite{cui2021learnable}, assuming that models trained on clean data embed their most discriminative features, constrains the logits from the robust model to make them similar to the model trained on natural data. This approach makes the robust model inherit the decision boundaries of the clean model, preserving high standard and robust accuracy.

\textbf{Low Temperature Distillation} (LTD)~\cite{chen2021ltd}, which uses previous distillation frameworks to generate labels, uses relatively low temperatures in the teacher model and employs different fixed temperatures for the teacher and student models. The main benefit of this mechanism is that the generated soft labels can be integrated into existing works without additional costs.

Recently, literature~\cite{yan2019robustness,haber2017stable,liu2020does} demonstrated that neural Ordinary Differential Equations (ODE) are naturally more robust to adversarial attacks than vanilla DNNs. Therefore, \textbf{Stable neural ODE for deFending against adversarial attacks} (SODEF)~\cite{kang2021stable} uses optimization formulation to force the extracted feature points to be within the vicinity of Lyapunov-stable equilibrium points, which suppresses the input perturbations.

\textbf{Self-COnsistent Robust Error} (SCORE)~\cite{pang2022robustness} employs local equivariance to describe the ideal behavior of a robust model, facilitating the reconciliation between robustness and accuracy while still dealing with worst-case uncertainty. This method was inspired by the discovery that the trade-off between adversarial and clean accuracy imposes a bias toward smoothness.

Analyzing the impact of activation shape on robustness, Dai \textit{et al.}~\cite{dai2022parameterizing} observes that activation has positive outputs on negative inputs, and a high finite curvature can improve robustness. Therefore, \textbf{Parametric Shifted Sigmoidal Linear Unit} (PSSiLU)~\cite{dai2022parameterizing} combines these properties and parameterized activation functions with adversarial training.

\subsection{Use of Supplementary Networks}

\textbf{MagNet}~\cite{Meng2017MagNetAT} considers two reasons for the misclassification of an adversarial example: 1) incapacity of the classifier to reject an adversarial example distant from the boundary; and 2) classifier generalizes poorly when the adversarial example is close to the boundary. MagNet considers multiple detectors trained based on the reconstruction error, detecting significantly perturbed examples and detecting slightly perturbed examples based on probability divergence.

\begin{figure}[!tb]
    \centering
    \includegraphics[width=.46\textwidth]{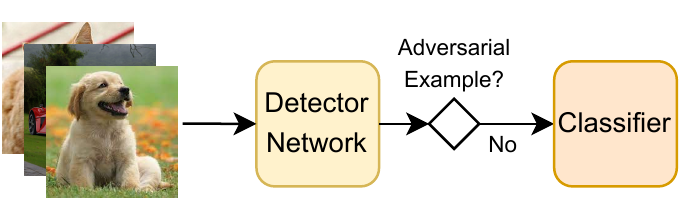}
    \caption{Schematic overview of the Use of Supplementary Networks. The Detector Network was previously trained to detect adversarial images and is included between the input images and the classifier. This network receives the input images and determines if these images are Adversarial or Not. If they are not, they are redirected to the Classifier; If they are, they are susceptible to Human evaluation.}
    \label{fig:supplementary_network}
\end{figure}

\textbf{Adversary Detection Network} (ADN)~\cite{Metzen2017Detecting} is a subnetwork that detects if the input example is adversarial or not, trained using adversarial images generated for a classification network which are classified as clean (0) or adversarial (1). Figure~\ref {fig:supplementary_network} displays a schematic overview of this network. However, this defense mechanism deeply correlates to the datasets and classification networks.

Xu \textit{et al.} found that the inclusion of \textbf{Feature Squeezing} (FS)~\cite{xu2017feature} is highly reliable in detecting adversarial examples by reducing the search space available for the adversary to modify. This method compares the predictions of a standard network with a squeezed one, detecting adversarial examples with high accuracy and having few false positives.

\textbf{High-level representation Guided Denoiser} (HGD)~\cite{Liao2018DefenseAA} uses the distance between original and adversarial images to guide an image denoiser and suppress the impact of adversarial examples. HGD uses a Denoising AutoEncoder~\cite{Vincent2008ExtractingAC} with additional lateral connections and considers the difference between the representations as the loss function at a specific layer that is activated by the normal and adversarial examples.

\textbf{Defense-GAN}~\cite{Samangouei2018DefenseGANPC} explores the use of GANs to effectively represent the set of original training examples, making this defense independent from the attack used. Defense-GAN considers the usage of Wasserstein GANs (WGANs)~\cite{Arjovsky2017WassersteinG} to learn the representation of the original data and denoise the adversarial examples, which start by minimizing the $l_2$ difference between the generator representation and the input image.
\textbf{Reverse Attacks}~\cite{Mao2021Reverse} can be applied to each attack during the testing phase, by finding the suitable additive perturbation to repair the adversarial example similar to the adversarial attacks, which is highly difficult due to the unknown original label.

\textbf{Embedding Regularized Classifier} (ER-Classifier)~\cite{Li_2021_ICCV} is composed of a classifier, an encoder, and a discriminator, which uses the encoder to generate code vectors by reducing the dimensional space of the inputs and the discriminator to separate these vectors from the ideal code vectors (sampled from a prior distribution). This technique allows pushing adversarial examples into the benign image data distribution, removing the adversarial perturbations.

\textbf{Class Activation Feature-based Denoiser} (CAFD)~\cite{Zhou2021NoiseCAM} is a self-supervised approach trained to remove the noise from adversarial examples, using a set of examples generated by the Class Activation Feature-based Attack (CAFA)~\cite{Zhou2021NoiseCAM}. This defense mechanism is trained to minimize the distance of the class activation features between the adversarial and natural examples, being robust to unseen attacks.

\textbf{Detector Graph} (DG)~\cite{Abusnaina2021ICCV} considers graphs to detect the adversarial examples by constructing a Latent Neighborhood Graph (LNG) for each original example and using Graph Neural Networks (GNNs)~\cite{Scarselli2009GNN} to exploit the relationship and distinguish between original and adversarial examples. This method maintains an additional reference dataset to retrieve the manifold information and uses embedding representation of image pixel values, making the defense robust to unseen attacks.

Images in the real world are represented in a continuous manner, yet machines can only store these images in discrete 2D arrays. \textbf{Local Implicit Image Function} (LIIF)~\cite{chen2021learning} takes an image coordinate and the deep features around this coordinate as inputs, predicting the corresponding RGB value. This method of pre-processing input images can filter adversarial images by reducing their perturbations, which are subsequently fed to a classifier.

\textbf{ADversarIal defenSe with local impliCit functiOns} (DISCO)~\cite{disco_2022} is an additional network to the classifier that removes adversarial perturbations using localized manifold projections, which receives an adversarial image and a query pixel location. This defense mechanism comprises an encoder that creates per-pixel deep features and a local implicit module that uses these features to predict the clean RGB value.

\begin{figure}[!tb]
    \centering
    \includegraphics[width=.46\textwidth]{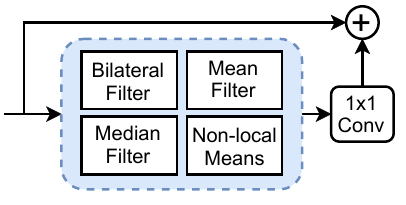}
    \caption{Overview of a Feature Denoising Block~\cite{Xie2018FDenoising}, which can be included in the intermediate layers to make networks more robust. This method is an example of Change Network Architecture.}
    \label{fig:feature_denoising}
\end{figure}

\subsection{Change Network Architecture}

To identify the type of layers and their order, Guo \textit{et al.}~\cite{Guo2020WhenNM} proposes the use of Neural Architecture Search (NAS) to identify the networks that are more robust to adversarial attacks, finding that densely connected patterns improve the robustness and adding convolution operations to direct connection edge is effective, combined to create the \textbf{RobNets}~\cite{Guo2020WhenNM}.

\textbf{Feature Denoising}~\cite{Xie2018FDenoising} intends to address this problem by applying feature-denoising operations, consisting of non-local means, bilateral, mean, median filters, followed by 1x1 Convolution and an identity skip connection, as illustrated in Figure~\ref{fig:feature_denoising}. These blocks are added to the intermediate layers of CNNs.

\textbf{Input Random}~\cite{Xie2017Mitigating} propose the addition of layers at the beginning of the classifier, consisting of 1) a random resizing layer, which resizes the width and height of the original image to a random width and height, and 2) a random padding layer, which pads zeros around the resized image in a random manner.

\textbf{Controlling Neural Level Sets} (CNLS)~\cite{atzmon2019controlling} uses samples obtained from the neural level sets and relates their positions to the network parameters, which allows modifying the decision boundaries of the network. The relation between position and parameters is achieved by constructing a sample network with an additional single fixed linear layer, which can incorporate the level set samples into a loss function.

\textbf{Sparse Transformation Layer} (STL)~\cite{sun2019adversarial}, included between the input image and the network first layer, transforms the received images into a low-dimensional quasi-natural image space, which approximates the natural image space and removes adversarial perturbations. This creates an attack-agnostic adversarial defense that gets the original and adversarial images closer.

Benz \textit{et al.}~\cite{Benz2020BNAdversarial} found that BN~\cite{Ioffe2015BatchNorm} and other normalization techniques make DNN more vulnerable to adversarial examples, suggesting the use of a framework that makes DNN more robust by learning \textbf{Robust Features} first and, then, Non-Robust Features (which are the ones learned when using BN).

\subsection{Perform Network Validation}

Most of the datasets store their images using the Joint Photographic Experts Group (JPEG)~\cite{good1994joint} compression, yet no one had evaluated the impact of this process on the network performance. Dziugaite \textit{et al.}~\cite{dziugaite2016study} (named as \textbf{JPG}) varies the magnitude of FGSM perturbations, discovering that smaller ones often reverse the drop in classification by a large extent and, when the perturbations increase in magnitude, this effect is nullified.

Regarding formal verification, a tool~\cite{Huang2017SafetyVO} for automatic \textbf{Safety Verification} of the decisions made during the classification process was created using Satisfiability Modulo Theory (SMT). This approach assumes that a decision is safe when, after applying transformations in the input, the model decision does not change. It is applied to every layer individually in the network, using a finite space of transformations.  

\textbf{DeepXplore}~\cite{Pei2017DeepXploreAW} is the first white-box framework to perform a wide test coverage, introducing the concepts of neuron coverage, which are parts of the DNN that are exercised by test inputs. DeepXplore uses multiple DNNs as cross-referencing oracles to avoid manual checking for each test input and inputs that trigger different behaviors and achieve high neuron coverage is a joint optimization problem solved by gradient-based search techniques.

\textbf{DeepGauge}~\cite{Ma2018DeepGaugeMT} intends to identify a testbed containing multi-faceted representations using a set of multi-granularity testing criteria. DeepGauge evaluates the resilience of DNNs using two different strategies, namely, primary function and corner-case behaviors, considering neuron- and layer-level coverage criteria. 

\textbf{Surprise Adequacy for Deep Learning Systems} (SADL)~\cite{Kim2019GuidingDL} is based on the behavior of DNN on the training data, by introducing the surprise of an input, which is the difference between the DNN behavior when given the input and the learned training data. The surprise of input is used as an adequacy criterion (Surprise Adequacy), which is used as a metric for the Surprise Coverage to ensure the input surprise range coverage.

The most recent data augmentation techniques, such as cutout~\cite{devries2017improved} and mixup~\cite{zhang2018mixup}, fail to prevent overfitting and, sometimes, make the model over-regularized, concluding that, to achieve substantial improvements, the combination of early stopping and semi-supervised data augmentation, \textbf{Overfit Reduction} (OR)~\cite{rice2020overfitting}, is the best method. 

When creating a model, multiple implementation details influence its performance; Pang \textit{et al.}~\cite{pang2020bag} is the first one to provide insights on how these details influence the model robustness, herein named as \textbf{Bag of Tricks} (BT). Some conclusions drawn from this study are: 1) The robustness of the models is significantly affected by weight decay; 2) Early stopping of the adversarial attacks may deteriorate worst-case robustness; and 3) Smooth activation benefits lower capacity models.

Overfitting is a known problem that affects model robustness; Rebuffi \textit{et al.}~\cite{rebuffi2021fixing} focuses on reducing this robust overfitting by using different data augmentation techniques. \textbf{Fixing Data Augmentation} (FDA)~\cite{rebuffi2021fixing} demonstrates that model weight averaging combined with data augmentation schemes can significantly increase robustness, which is enhanced when using spatial composition techniques.

Gowal \textit{et al.}~\cite{gowal2020uncovering} systematically studies the effect of multiple training losses, model sizes, activation functions, the addition of unlabeled data, and other aspects. The main conclusion drawn by this analysis is that larger models with Swish/SiLU~\cite{elfwing2018sigmoid} activation functions and model weight averaging can reliably achieve state-of-the-art results in robust accuracy.

\begin{figure}[!tb]
    \centering
    \includegraphics[width=.5\textwidth]{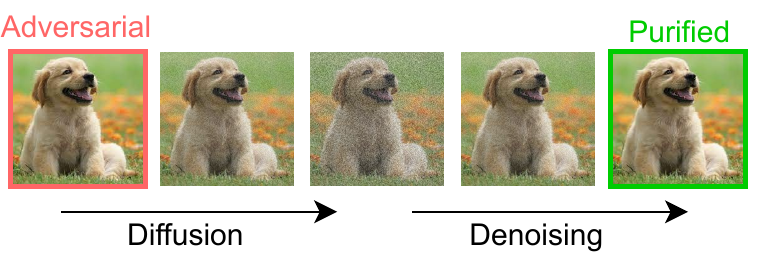}
    \caption{Overview of Adversarial Purification using Denoising Diffusion Probabilistic Models, adapted from~\cite{ho2020denoising}. The diffusion process is applied to an adversarial image, consisting of adding noise for a certain number of steps. During the denoising procedure, this noise is iteratively removed by the same amount of steps, resulting in a purified image (without perturbations).}
    \label{fig:adversarial_purification}
\end{figure}

\subsection{Adversarial Purification}

Adversarial Purification consists of defense mechanisms that remove adversarial perturbations using a generative model. \textbf{Improving Robustness Using Generated Data} (IRUGD)~\cite{gowal2021improving} explores how generative models trained on the original images can be leveraged to increase the size of the original datasets. Through extensive experiments, they concluded that Denoising Diffusion Probabilistic Models (DDPM)~\cite{ho2020denoising}, a progression of diffusion probabilistic models~\cite{sohl2015deep}, is the model that more closely resembles real data. Figure~\ref{fig:adversarial_purification} presents the main idea behind the DDPM process.

Due to the great results in image synthesis displayed by the DDPM, Sehwag \textit{et al.}~\cite{sehwag2021robust} (\textbf{Proxy}) uses proxy distributions to significantly improve the performance of adversarial training by generating additional examples, demonstrating that the best generative models for proxy distribution are DDPM.

Inspired by previous works on adversarial purification~\cite{shi2021online,yoon2021adversarial}, \textbf{DiffPure}~\cite{nie2022diffusion} uses DDPM for adversarial purification, receiving as input an adversarial example and recovering the clean image through a reverse generative process. Since this discovery, multiple improvements regarding the use of DDPM for Adversarial Purification have been studied. \textbf{Guided Diffusion Model for Adversarial Purification} (GDMAP)~\cite{wu2022guided} receives as initial input pure Gaussian noise and gradually denoises it with guidance to an adversarial image.

\textbf{DensePure}~\cite{xiao2022densepure} employs iterative denoising to an input image, with different random seeds, to get multiple reversed samples, which are given to the classifier and the final prediction is based on majority voting. Furthermore, Wang \textit{et al.}~\cite{wang2023better} uses the most recent diffusion models~\cite{karras2022elucidating} to demonstrate that diffusion models with higher efficiency and image quality directly translate into better robust accuracy.

\section{Adversarial Effects on Vision Transformers}

Like CNNs~\cite{Krizhevsky2012CNNs}, the ViTs are also susceptible to adversarial perturbations that alter a patch in an image~\cite{Naseer2021Properties}, and ViTs demonstrate higher robustness, almost double, compared with ResNet-50~\cite{He2015ResNet}. 

To further evaluate the \textbf{robustness of ViT} to adversarial examples, Mahmood \textit{et al.}~\cite{Mahmood2021RobustnessViT} used multiple adversarial attacks in CNNs, namely FGSM, PGD, MIM, C\&W, and MI-FGSM. The ViT has increased robustness (compared with ResNet) for the first four attacks and has no resilience to the C\&W and MI-FGSM attacks. Additionally, to complement the results obtained from the performance of ViTs, an extensive study~\cite{Aldahdooh2021RevealViT} using feature maps, attention maps, and Gradient-weighted Class Activation Mapping (Grad-CAM)~\cite{Selvaraju2017GradCAM} intends to explain this performance visually. 

The transferability of adversarial examples from CNNs to ViTs was also evaluated, suggesting that the examples from CNNs do not instantly transfer to ViTs~\cite{Mahmood2021RobustnessViT}. Furthermore, \textbf{Self-Attention blended Gradient Attack} (SAGA)~\cite{Mahmood2021RobustnessViT} was proposed to misclassify both ViTs and CNNs. The \textbf{Pay No Attention} (PNA)~\cite{Wei2021TransferabilityViT} attack, which ignores the gradients of attention, and the \textbf{PatchOut}~\cite{Wei2021TransferabilityViT} attack, which randomly samples subsets of patches, demonstrate high transferability.

To detect adversarial examples that might affect the ViTs, \textbf{PatchVeto}~\cite{Huang2021Defense} uses different transformers with different attention masks that output the encoding of the class. An image is considered valid if all transformers reach a consensus in the voted class, overall the masked predictions (provided by masked transformers).

\textbf{Smoothed ViTs}~\cite{Salman2021SmoothViT} perform preprocessing techniques to the images before feeding them into the ViT, by generating image ablations (images composed of only one column of the original image, and the remaining columns are black), which are converted into tokens, and droping the fully masked tokens. The remaining tokens are fed into a ViT, which predicts a class for each ablation, and the class with the most predictions of overall ablations is considered the correct one.

Bai \textit{et al.}~\cite{bai2021transformers} demonstrates that ViTs and CNNs are being unfairly evaluated because they do not have the same training details. Therefore, this work provides a fair and in-depth comparison between ViTs and CNNs, indicating that ViTs are as vulnerable to adversarial perturbations as CNNs.

\textbf{Architecture-oriented Transferable Attacking} (ATA)~\cite{wang2022generating} is a framework that generates transferable adversarial examples by considering the common characteristics among different ViT architectures, such as self-attention and image-embedding. Specifically, it discovers the most attentional patch-wise regions significantly influencing the model decision and searches pixel-wise attacking positions using sensitive embedding perturbation.

\textbf{Patch-fool}~\cite{fu2022patchfool} explores the perturbations that turn ViTs more vulnerable learners than CNNs, proposing a dedicated attack framework that fools the self-attention mechanism by attacking a single patch with multiple attention-aware optimization techniques. This attack mechanism demonstrates, for the first time, that ViTs can be more vulnerable than CNNs if attacked with proper techniques.

Gu \textit{et al.}~\cite{gu2022vision} evaluates the robustness of ViT to patch-wise perturbations, concluding that these models are more robust to naturally corrupted patches than CNNs while being more vulnerable to adversarially generated ones. Inspired by the observed results, the authors propose a simple \textbf{Temperature Scaling} based method that improves the robustness of ViTs.

As previously observed for CNNs, improving the robust accuracy sacrifices the standard accuracy of ViTs, which may limit their applicability in the real context. \textbf{Derandomized Smoothing}~\cite{chen2022towards} uses a progressive smoothed image modeling task to train the ViTs, making them capture the more discriminating local context while preserving global semantic information, improving both robust and standard accuracy.

\textbf{VeinGuard}~\cite{li2023transformer} is a defense framework that helps ViTs be more robust against adversarial palm-vein image attacks, with practical applicability in the real world. Namely, VeinGuard is composed of a local transformer-based GAN that learns the distribution of unperturbed vein images and a purifier that automatically removes a variety of adversarial perturbations.

\section{Datasets}
\label{sec:sota_datasets}

\subsection{MNIST and F-MNIST}

One of the most used datasets is the \textbf{MNIST}~\cite{lecun1998gradient} dataset, which contains images of handwritten digits collected from approximately 250 writers in shades of black and white, withdrawn from two different databases. This dataset is divided into training and test sets, with the first one containing 60,000 examples and a second one containing 10,000 examples. 

Xiao \textit{et al.} propose the creation of the \textbf{Fashion-MNIST}~\cite{xiao2017fashion} dataset by using figures from a fashion website, which has a total size of 70,000 images, contains ten classes, uses greyscale images, and each image has a size of 28x28. The Fashion-MNIST dataset is divided into train and test sets, containing 60,000 and 10,000 examples, respectively.

\begin{figure}[!tb]
    \centering
    \includegraphics[width=0.48\textwidth]{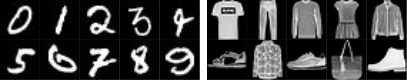}
    \caption{Images withdrew from the MNIST dataset~\cite{lecun1998gradient} in the first five columns and from the Fashion-MNIST dataset~\cite{xiao2017fashion} in the last five columns.  The images were resized for better visualization.}
    \label{fig:MNIST_FMNIST_images}
\end{figure}

Fig.~\ref{fig:MNIST_FMNIST_images} displays the 10 digits (from 0 to 9) from the MNIST dataset in the first five columns and the 10 fashion objects from Fashion-MNIST dataset in the last five columns. MNIST is one of the most widely studied datasets in the earlier works of adversarial examples, with defense mechanisms already displaying high robustness on this dataset. The same does not apply to Fashion-MNIST, which has not been as widely studied, despite having similar characteristics to MNIST.

\begin{figure}[!tb]
    \centering
    \includegraphics[width=0.48\textwidth]{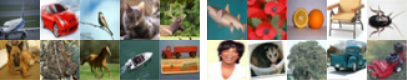}
    \caption{Images withdrew from the CIFAR-10 dataset~\cite{krizhevsky2009learning} in the first five columns and from the CIFAR-100 dataset~\cite{krizhevsky2009learning} in the last five columns. The images were resized for better visualization.}
    \label{fig:CIFAR10_CIFAR100_images}
\end{figure}

\subsection{CIFAR-10 and CIFAR-100}
Another widely studied dataset is the CIFAR-10, which, in conjunction with the CIFAR-100 dataset, are subsets from a vast database containing 80 million tiny images~\cite{torralba2008tiny}, 32x32, and three color channels 75,062 different classes.

\textbf{CIFAR-10}~\cite{krizhevsky2009learning} contains only ten classes from this large database, with 6,000 images for each class, distributed into 50,000 training images and 10,000 test images. This dataset considers different objects, namely, animals and vehicles, usually found in different environments.

\textbf{CIFAR-100}~\cite{krizhevsky2009learning} contains 100 classes with only 600 images for each one with the same size and amount of color channels as the CIFAR-10 dataset. CIFAR-100 groups its 100 classes into 20 superclasses, located in different contexts/environments, making this dataset much harder to achieve high results.

Examples from the CIFAR-10 dataset are shown in Fig.~\ref{fig:CIFAR10_CIFAR100_images} in the first five columns, and the remaining columns display examples of the superclasses from CIFAR-100. Due to the unsatisfactory results demonstrated by models trained on CIFAR-10, the CIFAR-100 dataset has not been included in most studies under the context of adversarial examples, suggesting that solving the issue of adversarial-perturbed images is still at its inception.

\begin{figure}[!tb]
    \centering
    \includegraphics[width=0.48\textwidth]{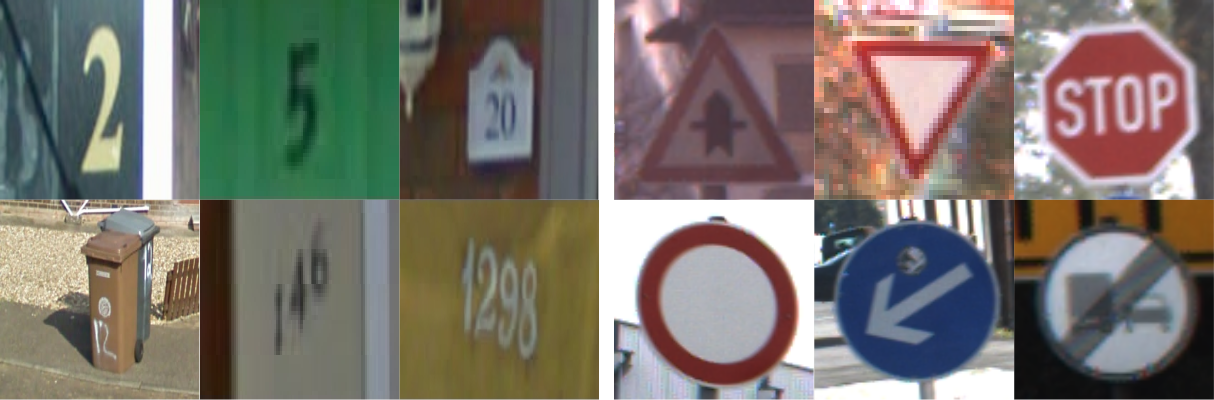}
    \caption{Images withdrew from the Street View House Numbers dataset~\cite{netzer2011svhn} in the first three columns and from the German Traffic Sign Recognition Benchmark dataset~\cite{stallkamp2012traffic} in the last three columns. The images were resized for better visualization.}
    \label{fig:SVHN_GTSRB_images}
\end{figure}

\begin{figure*}[!tb]
    \centering
    \includegraphics[width=0.9\textwidth]{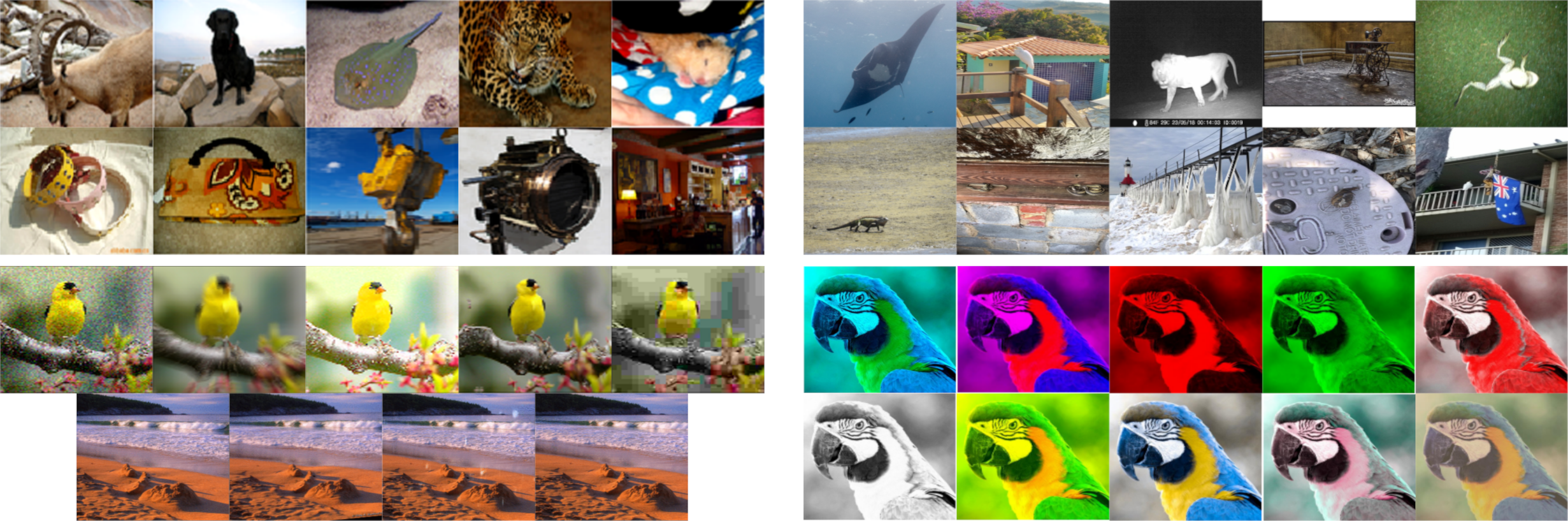}
    \caption{Images withdrew from the ImageNet dataset~\cite{russakovsky2015imagenet} in the top left, from the ImageNet-A dataset~\cite{Hendrycks_2021_CVPR} in the top right, from the ImageNet-C and ImageNet-P datasets~\cite{hendrycks2018benchmarking} in the bottom left, and ImageNet-COLORDISTORT~\cite{De2021ImageNetCD} in the bottom right. The images were resized for better visualization.}
    \label{fig:ImageNet_variants_images}
\end{figure*}

\subsection{Street View Datasets}

The \textbf{Street View House Numbers} (SVHN)~\cite{netzer2011svhn} dataset provides the same challenge as MNIST: identifying which digits are present in a colored image, containing ten classes, 0 to 9 digits, and an image size of 32x32 centered around a single character, with multiple digits in a single image. Regarding the dataset size, it has 630,420 digit images, but only 73,257 images are used for training, 26,032 images are used for testing, and the remaining 531,131 images can be used as additional training data.

\textbf{German Traffic Sign Recognition Benchmark} (GTSRB)~\cite{stallkamp2012traffic} is a dataset containing 43 classes of different traffic signs, has 50,000 images, and demonstrates realistic scenarios. The dataset has 51,840 images, whose size varies from 15x15 to 222x193, divided into training, validation, and test sets with 50\%, 25\%, and 25\%, respectively, of the total images.

The difficulties associated with the SVHN dataset are displayed in the first three rows of Fig.~\ref{fig:SVHN_GTSRB_images}, showing unique digits that occupy the whole image and multiple digits on different backgrounds. Furthermore, the same figure presents the different types of traffic signs in the GTSRB dataset, such as prohibition, warning, mandatory, and end of prohibition.

\subsection{ImageNet and Variants}
\textbf{ImageNet}~\cite{russakovsky2015imagenet} is one of the largest datasets for object recognition, containing 1,461,406 colored images and 1,000 classes, with images being resized to 224x224. This dataset collected photographs from Flickr, and other search engines, divided into 1.2 million training images, 50,000 validation images, and 100,000 test images.

A possible alternative to ImageNet, when the dataset size is an important factor, is called \textbf{Tiny ImageNet}~\cite{le2015tiny}, a subset of ImageNet that contains fewer classes and images. This dataset contains only 200 classes (from the 1,000 classes in ImageNet), 100,000 training images, 10,000 validation images, and 10,000 test images. These classes include animals, vehicles, household items, insects, and clothing, considering the variety of contexts/environments that these objects can be found. Their images have a size of 64x64 and are colored. 

\textbf{ImageNet-A}~\cite{Hendrycks_2021_CVPR} is a subset of ImageNet, containing only 200 classes from the 1,000 classes, covering the broadest categories in ImageNet. ImageNet-A is a dataset composed of real-world adversarially filtered images, which were obtained by deleting the correctly predicted images by ResNet-50 classifiers. Despite ImageNet-A being based on the deficiency of ResNet-50, it also demonstrates transferability to unseen models, making this dataset suitable for evaluating the robustness of multiple classifiers.

Two additional benchmarks, \textbf{ImageNet-C}~\cite{hendrycks2018benchmarking} and \textbf{ImageNet-P}~\cite{hendrycks2018benchmarking}, were designed to evaluate the robustness of DNNs. The ImageNet-C standardizes and expands the corruption robustness topic, consisting of 75 corruptions applied to each image in the ImageNet validation set. ImageNet-P applies distortions to the images, though it differs from ImageNet-C because it contains perturbation sequences using only ten common perturbations.

\begin{table}[!tb]
    \centering
    \caption{Relevant characteristics to the context of adversarial examples of the state-of-the-art datasets. \textbf{\#Classes} means the number of classes in the dataset. Empty \textbf{Color} column means that the images in that dataset use greyscale or black and white shades. Datasets with $*$ are only used for testing purposes.}
    \begin{tabular}{@{}cccccc@{}}
        \hline
        \textbf{Dataset} & \textbf{Size} & \textbf{\#Classes} & \textbf{Classes} & \textbf{Color} \\
        \hline\hline
        
        \multirow{2}{*}{MNIST} & \multirow{2}{*}{70,000} & \multirow{2}{*}{10} & \multirow{2}{*}{Digits} & \\
         & & & & \\
         
        \hline
        \multirow{2}{*}{Fashion-MNIST} & \multirow{2}{*}{70,000} & \multirow{2}{*}{10} & \multirow{2}{*}{Clothing} & \\
         & & & & \\
          
        \hline
        \multirow{2}{*}{CIFAR-10} & \multirow{2}{*}{60,000} & \multirow{2}{*}{10} &  Animals & \multirow{2}{*}{\checkmark} \\
         & & & Vehicles & \\
         
        \hline
        \multirow{2}{*}{SVHN} & \multirow{2}{*}{630,420} & \multirow{2}{*}{10} & \multirow{2}{*}{Digits} & \multirow{2}{*}{\checkmark} \\
         & & & & \\
         
        \hline
        \multirow{2}{*}{GTSRB} & \multirow{2}{*}{51,840} & \multirow{2}{*}{43} & \multirow{2}{*}{Traffic Signs} & \multirow{2}{*}{\checkmark} \\
         & & & & \\
         
        \hline
        \multirow{2}{*}{CIFAR-100} & \multirow{2}{*}{60,000} & \multirow{2}{*}{100} & Household Items & \multirow{2}{*}{\checkmark} \\
         & & & Outdoor Scenes & \\
        
        \hline
        \multirow{2}{*}{Tiny ImageNet} & \multirow{2}{*}{120,000} & \multirow{2}{*}{200} & Animals & \multirow{2}{*}{\checkmark} \\
         & & & Household Items & \\
        
        \hline
        \multirow{2}{*}{ImageNet-A$^*$} & \multirow{2}{*}{7,500} & \multirow{2}{*}{200} & Vehicles & \multirow{2}{*}{\checkmark} \\
         & & & Food & \\
        
        \hline
        \multirow{2}{*}{ImageNet-C$^*$} & \multirow{2}{*}{3,750,000} & \multirow{2}{*}{200} & Vehicles & \multirow{2}{*}{\checkmark} \\
         & & & Food & \\
        
        \hline
        \multirow{2}{*}{ImageNet-P$^*$} & \multirow{2}{*}{15,000,000} & \multirow{2}{*}{200} & Vehicles & \multirow{2}{*}{\checkmark} \\
         & & & Food & \\
        
        \hline
        \multirow{2}{*}{ImageNet} & \multirow{2}{*}{1,431,167} & \multirow{2}{*}{1,000} & Vehicles & \multirow{2}{*}{\checkmark} \\
         & & & Electronic devices & \\
        
        \hline
        \multirow{2}{*}{ImageNet-CD$^*$} & \multirow{2}{*}{736,515} & \multirow{2}{*}{1,000} & Vehicles & \multirow{2}{*}{\checkmark} \\
         & & & Electronic devices & \\
        \hline
    \end{tabular}
    \label{tab:datasets_info}
\end{table}

Another benchmark to evaluate the model generalization capability is the \textbf{ImageNet-COLORDISTORT} (ImageNet-CD)~\cite{De2021ImageNetCD}, which considers multiple distortions in the color of an image using different color space representations. This dataset contains the 1,000 classes from ImageNet, removing images without color channels, and the same image considers multiple color distortions under the Red Green Blue (RGB), Hue-Saturation-Value (HSV), CIELAB, and YCbCr color spaces considered common transformations used in image processing.

It is possible to observe a set of images withdrawn from ImageNet in the top left of Fig.~\ref{fig:ImageNet_variants_images}. Additionally, some images misclassified by multiple classifiers (ImageNet-A) are shown in the top right of the same figure. The bottom represents the ImageNet with common corruptions and perturbations and is manipulated by multiple image techniques on the left and right, respectively. Table~\ref{tab:datasets_info} summarizes the main characteristics of the datasets presented throughout this section.

\begin{table}[!tb]
    \centering
    \caption{Accuracy comparison of different defense mechanisms on CIFAR-10 under PGD attack, $l_{\infty}$ and $\epsilon = 8/255$. Clean and Robust refers to accuracy Without and With Adversarial Attacks, respectively. Defenses with ``-'' on clean accuracy do not have a clean accuracy reported.}
    \begin{tabular}{l|c|c|c|c}
         \multirow{2}{*}{Defense Method} & \multirow{2}{*}{Year} & \multirow{2}{*}{Architecture} & \multicolumn{2}{c}{Accuracy} \\
         & & & Clean & Robust \\
        \hline\hline

        BPFC~\cite{Addepali2020BPFC} & 2020 & ResNet-18 & 82.4 & 34.4 \\
        SNS~\cite{Zhang2021InterpretingAI} & 2021 & VGG-16 & 86.0 & 39.6 \\
        AT-MIFGSM~\cite{Kurakin2017AdversarialML} & 2017 & Inception v3 & 85.3 & 45.9 \\
        AT-PGD~\cite{Madry2018TowardsDL} & 2018 & ResNet-18 & 87.3 & 47.0 \\
        RobNets~\cite{Guo2020WhenNM} & 2020 & RobNet-free & 82.8 & 52.6 \\
        HGD~\cite{Liao2018DefenseAA} & 2018 & DUNET & 92.4 & 53.1 \\
        RSLAD~\cite{Zi2021RSLAD} & 2021 & ResNet-18 & 83.4 & 54.2 \\
        MART~\cite{wang2020improving} & 2020 & WRN-28-10  & 83.1 & 55.6 \\
        TRADES~\cite{zhang2019theoretically} & 2019 & WRN-34-10 & 84.9 & 56.4 \\
        BagT~\cite{pang2020bag} & 2020 & WRN-34-10 & - & 56.4 \\
        RO~\cite{rice2020overfitting} & 2020 & ResNet-18 & - & 56.8 \\
        DOA~\cite{wu2019defending} & 2019 & VGGFace & 93.6 & 61.0 \\
        AWP~\cite{wu2020adversarial} & 2020 & WRN-28-10 & - & 63.6 \\
        FS~\cite{Zhang2019DefenseAA} & 2019 & WRN-28-10 & 90.0 & 68.6 \\
        CAFD~\cite{Zhou2021NoiseCAM} & 2021 & DUNET & 91.1 & 87.2 \\      
        \hline
    \end{tabular}
    \label{tab:comparison_defenses_PGD_Cifar10}
\end{table}

\section{Metrics and State-of-the-art Results}
\label{sec:sota_metrics_results}

\subsection{Evaluation Metrics}

Due to the nature of adversarial examples, they need specific metrics to be correctly evaluated and constructed. Following this direction, multiple works have been proposing different metrics that calculate the percentage of adversarial examples that make a model misclassify (fooling rate), measure the amount of perturbation made in an image (destruction rate), and calculate the model robustness to adversarial examples (average robustness).

\subsubsection{Accuracy}
This metric measures the number of samples that are correctly predicted by the model, which is defined as:
\begin{equation}
    \text{accuracy} = \frac{TP + TN}{TP + TN + FP + FN},
\end{equation}
where $TP$ refers to True Positive, $TN$ to True Negative, $FP$ to False Positive, and $FN$ to False Negative. The True Positive and True Negative are the samples whose network prediction is the same as the label (correct), and the False Positive and False Negative are the samples whose network prediction differs from the label (incorrect). When considering original images, this metric is denominated as \textit{Clean Accuracy} and, when using adversarial images, is named as \textit{Robust Accuracy}.

\subsubsection{Fooling Rate}

After being perturbed to change the classifier label, the fooling rate $FR$~\cite{huan2020data} was proposed to calculate the percentage of images.

\subsubsection{Average Robustness}

To objectively evaluate the robustness to adversarial perturbations of a classifier $f$, the average robustness $\hat{p}_{\text{adv}}(f)$ is defined as~\cite{MoosaviDezfooli2016DeepFoolAS}:

\begin{equation}
    \hat{p}_{\text{adv}}(f) = \frac{1}{\mathcal{D}} \sum_{x \in \mathcal{D}} \frac{\| \hat{r}(x) \|_2}{\| x \|_2},
\end{equation}
where $\hat{r}(x)$ is the estimated minimal perturbation obtained using the attack, and $\mathcal{D}$ denotes the test set.

\begin{table}[!tb]
    \centering
    \caption{Accuracy comparison of different defense mechanisms on CIFAR-10 under Auto-Attack attack, $l_{\infty}$ and $\epsilon = 8/255$ . Clean and Robust refers to accuracy Without and With Adversarial Attacks, respectively.}
    \begin{tabular}{l|c|c|c|c}
         \multirow{2}{*}{Architecture} & \multirow{2}{*}{Defense Method} & \multirow{2}{*}{Year} & \multicolumn{2}{c}{Accuracy} \\
         & & & Clean & Robust \\
        \hline\hline

        \multirow{23}{*}{WRN28-10} 
         & Input Random~\cite{Xie2017Mitigating} & 2017 & 94.3 & 8.6 \\
         & BAT~\cite{wang2019bilateral} & 2019 & 92.8 & 29.4 \\
         & FS~\cite{Zhang2019DefenseAA} & 2019 & 90.0 & 36.6 \\
         & Jpeg~\cite{dziugaite2016study} & 2016 & 83.9 & 50.7 \\
         & Pretrain~\cite{hendrycks2019using} & 2019 & 87.1 & 54.9 \\
         & UAT~\cite{alayrac2019labels} & 2019 & 86.5 & 56.0 \\
         & MART~\cite{wang2020improving} & 2020 & 87.5 & 56.3 \\
         & HYDRA~\cite{sehwag2020hydra} & 2020 & 89.0 & 57.1 \\
         & RST~\cite{carmon2019unlabeled} & 2019 & 89.7 & 59.5 \\
         & GI-AT~\cite{zhang2020geometry} & 2020 & 89.4 & 59.6 \\
         & Proxy~\cite{sehwag2021robust} & 2021 & 89.5 & 59.7 \\
         & AWP~\cite{wu2020adversarial} & 2020 & 88.3 & 60.0 \\
         & FDA~\cite{rebuffi2021fixing} & 2021 & 87.3 & 60.8 \\
         & HAT~\cite{rade2021helper} & 2021 & 88.2 & 61.0 \\
         & SCORE~\cite{pang2022robustness} & 2022 & 88.6 & 61.0 \\
         & PSSiLU~\cite{dai2022parameterizing} & 2022 & 87.0 & 61.6 \\
         & Gowal \textit{et al.}~\cite{gowal2020uncovering} & 2020 & 89.5 & 62.8 \\
         & IRUGD~\cite{gowal2021improving} & 2021 & 87.5 & 63.4 \\
         & Wang \textit{et al.}~\cite{wang2023better} & 2023 & 92.4 & 67.3 \\
         & STL~\cite{sun2019adversarial} & 2019 & 82.2 & 67.9 \\
         & DISCO~\cite{disco_2022} & 2022 & 89.3 & 85.6 \\
         
        \hline

        \multirow{16}{*}{WRN34-10} 
        & Free-AT~\cite{shafahi2019adversarial} & 2019 & 86.1 & 41.5 \\
        & AT-PGD~\cite{Madry2018TowardsDL} & 2018 & 87.1 & 44.0 \\
        & YOPO~\cite{zhang2019you} & 2019 & 87.2 & 44.8 \\
        & TLA~\cite{mao2019metric} & 2019 & 86.2 & 47.4 \\
        & LAT~\cite{kumari2019harnessing} & 2019 & 87.8 & 49.1 \\
        & SAT~\cite{sitawarin2020improving} & 2020 & 86.8 & 50.7 \\
        & FAT~\cite{chen2022efficient} & 2022 & 85.3 & 51.1 \\ 
        & LBGAT~\cite{cui2021learnable} & 2021 & 88.2 & 52.3 \\ 
        & TRADES~\cite{zhang2019theoretically} & 2019 & 84.9 & 53.1 \\ 
        & SAT~\cite{huang2020self} & 2020 & 83.5 & 53.3 \\ 
        & Friend-AT~\cite{zhang2020attacks} & 2020 & 84.5 & 55.5 \\ 
        & AWP~\cite{wu2020adversarial} & 2020 & 85.4 & 56.2 \\ 
        & LTD~\cite{chen2021ltd} & 2021 & 85.2 & 56.9 \\
        & OA-AT~\cite{addepalli2021towards} & 2021 & 85.3 & 58.0 \\ 
        & Proxy~\cite{sehwag2021robust} & 2022 & 86.7 & 60.3 \\
        & HAT~\cite{rade2021helper} & 2021 & 91.5 & 62.8 \\

        \hline

        \multirow{6}{*}{WRN-70-16} 
        & SCORE~\cite{pang2022robustness} & 2022 & 89.0 & 63.4 \\
        & IRUGD~\cite{gowal2021improving} & 2021 & 91.1 & 65.9 \\
        & Gowal \textit{et al.}~\cite{gowal2020uncovering} & 2020 & 88.7 & 66.1 \\
        & FDA~\cite{rebuffi2021fixing} & 2021 & 92.2 & 66.6 \\
        & Wang \textit{et al.}~\cite{wang2023better} & 2023 & 93.3 & 70.7 \\
        & SODEF~\cite{kang2021stable} & 2021 & 93.7 & 71.3 \\

        \hline
    \end{tabular}
    \label{tab:comparison_defenses_AA_CIFAR10}
\end{table}

\subsubsection{Destruction Rate}

To evaluate the impact of arbitrary transformations on adversarial images, the notion of destruction rate $d$ is introduced and formally defined as~\cite{Kurakin2017AdversarialEI}:

\begin{equation}
    d = \frac{ \sum^{n}_{k=1} C( \textbf{X}^k, y^k_{\text{true}} ) \neg C( \textbf{X}^k_{\text{adv}}, y^k_{\text{true}} ) C( T(\textbf{X}^k_{\text{adv}}), y^k_{\text{true}} )}{ \sum^{n}_{k=1} ( \textbf{X}^k, y^k_{\text{true}} ) C( \textbf{X}^k_{\text{adv}}, y^k_{\text{true}} ) },
\end{equation}
where $n$ is the number of images, $\mathbf{X}^k$ is the original image from the dataset, $y^k_{\text{true}}$ is the true class of this image, $\mathbf{X}^k_{\text{adv}}$ is the adversarial image corresponding to that image, and $T$ is an arbitrary image transformation. $\neg C( \mathbf{X}^k_{\text{adv}}, y^k_{\text{true}} )$ is defined as the binary negation of $C( \mathbf{X}^k_{\text{adv}}, y^k_{\text{true}} )$. Finally, the function $C(\mathbf{X}, y)$ is defined as~\cite{Kurakin2017AdversarialEI}:

\begin{equation}
    C(\mathbf{X}, y) = \begin{cases} 1, & \textrm{if image }\textbf{X}\textrm{ is classified as } y; \\ 0, & \textrm{otherwise}. \end{cases}
\end{equation}

\subsection{Defense Mechanisms Robustness}

The metric used to evaluate models is accuracy, which evaluates the results on both original (\textbf{Clean Accuracy}) and adversarially perturbed (\textbf{Robust Accuracy}) datasets. One of the earliest and strongest adversarial attacks proposed was PGD, which was used by multiple defenses to evaluate their robustness. Table~\ref{tab:comparison_defenses_PGD_Cifar10} displays defenses evaluated on CIFAR-10 under multiple steps PGD attack, ordered by increasing robustness. For the PGD attack, the best performing defenses are from approaches that use supplementary networks (CAFD) or modify the training process (FS and AWP). Overall, Wide ResNets~\cite{zagoruyko2016wide} have better robust accuracy, due to high-capacity networks exhibiting greater adversarial robustness~\cite{Kurakin2017AdversarialML, Madry2018TowardsDL}, suggesting the usage of these networks in future developments of adversarial attacks and defenses.

To assess the robustness of defenses for white and black-box settings, Auto-Attack has gained increased interest over PGD in recent works. Tables~\ref{tab:comparison_defenses_AA_CIFAR10},~\ref{tab:comparison_defenses_AA_CIFAR100}, and~\ref{tab:comparison_defenses_AA_ImageNet} present a set of defenses that are evaluated under Auto-Attack, on CIFAR-10, CIFAR-100, and ImageNet, respectively, ordered by increasing Robust Accuracy. The most used networks are Wide ResNets with different sizes, with the biggest Wide ResNet displaying better results overall, and the most resilient defense derives from the use of supplementary networks (DISCO), followed by modifying the train process (SODEF) and changing network architecture (STL). The results suggest that the inclusion of additional components to sanitize inputs of the targeted model (use of supplementary networks) is the most resilient approach for model robustness in white and black-box settings. 
The updated results for defenses under Auto-Attack can be found on the RobustBench~\cite{croce2021robustbench} website.

\begin{table}[!tb]
    \centering
    \caption{Accuracy comparison of different defense mechanisms on CIFAR-100 under Auto-Attack attack, $l_{\infty}$ and $\epsilon = 8/255$ . Clean and Robust refers to accuracy Without and With Adversarial Attacks, respectively.}
    \begin{tabular}{l|c|c|c|c}
         \multirow{2}{*}{Architecture} & \multirow{2}{*}{Defense Method} & \multirow{2}{*}{Year} & \multicolumn{2}{c}{Accuracy} \\
         & & & Clean & Robust \\
        \hline\hline

        \multirow{10}{*}{WRN28-10} 
         & Input Random~\cite{Xie2017Mitigating} & 2017 & 73.6 & 3.3 \\
         & LIIF~\cite{chen2021learning} & 2021 & 80.3 & 3.4 \\
         & Bit Reduction~\cite{xu2017feature} & 2017 & 76.9 & 3.8 \\
         & Pretrain~\cite{hendrycks2019using} & 2019 & 59.2 & 28.4 \\
         & SCORE~\cite{pang2022robustness} & 2022 & 63.7 & 31.1 \\
         & FDA~\cite{rebuffi2021fixing} & 2021 & 62.4 & 32.1 \\
         & Wang \textit{et al.}~\cite{wang2023better} & 2023 & 78.6 & 38.8 \\
         & Jpeg~\cite{dziugaite2016study} & 2016 & 61.9 & 39.6 \\
         & STL~\cite{sun2019adversarial} & 2019 & 67.4 & 46.1 \\
         & DISCO~\cite{disco_2022} & 2022 & 72.1 & 67.9 \\
         
        \hline

        \multirow{7}{*}{WRN34-10} 
        & SAT~\cite{sitawarin2020improving} & 2020 & 62.8 & 24.6 \\
        & AWP~\cite{wu2020adversarial} & 2020 & 60.4 & 28.9 \\ 
        & LBGAT~\cite{cui2021learnable} & 2021 & 60.6 & 29.3 \\ 
        & OA-AT~\cite{addepalli2021towards} & 2021 & 65.7 & 30.4 \\ 
        & LTD~\cite{chen2021ltd} & 2021 & 64.1 & 30.6 \\ 
        & Proxy~\cite{sehwag2021robust} & 2022 & 65.9 & 31.2 \\ 
        & DISCO~\cite{disco_2022} & 2022 & 71.6 & 69.0 \\

        \hline

        \multirow{4}{*}{WRN-70-16} 
         & SCORE~\cite{pang2022robustness} & 2022 & 65.6 & 33.1 \\
         & FDA~\cite{rebuffi2021fixing} & 2021 & 63.6 & 34.6 \\
         & Gowal \textit{et al.}~\cite{gowal2020uncovering} & 2020 & 69.2 & 36.9 \\
         & Wang \textit{et al.}~\cite{wang2023better} & 2023 & 75.2 & 42.7 \\
    
        \hline
    \end{tabular}
    \label{tab:comparison_defenses_AA_CIFAR100}
\end{table}

\section{Future Directions}
\label{sec:fut_dir}

Following the \textit{de facto} standards adopted by the literature, we suggest that future proposals of \textbf{defense mechanisms should be evaluated on Auto-Attack}, using the \textbf{robust accuracy} as a metric for comparison purposes. The adversarial defense that demonstrates better results is \textbf{Adversarial Training}, which should be a \textbf{requirement when evaluating attacks and defenses}.

\begin{table}[!tb]
    \centering
    \caption{Accuracy comparison of different defense mechanisms on ImageNet under Auto-Attack attack, $l_{\infty}$ and $\epsilon = 4/255$. Clean and Robust refers to accuracy Without and With Adversarial Attacks, respectively.}
    \begin{tabular}{l|c|c|c|c}
         \multirow{2}{*}{Architecture} & \multirow{2}{*}{Defense Method} & \multirow{2}{*}{Year} & \multicolumn{2}{c}{Accuracy} \\
         & & & Clean & Robust \\
        \hline\hline

        \multirow{6}{*}{ResNet-18} 
         & Bit Reduction~\cite{xu2017feature} & 2017 & 67.6 & 4.0 \\
         & Jpeg~\cite{dziugaite2016study} &  2016 & 67.2 & 13.1 \\
         & Input Random~\cite{Xie2017Mitigating} & 2017 & 64.0 & 17.8 \\
         & Salman \textit{et al.}~\cite{salman2020adversarially} & 2020 & 52.9 & 25.3 \\
         & STL~\cite{sun2019adversarial} & 2019 & 65.6 & 32.9 \\
         & DISCO~\cite{disco_2022} & 2022 & 68.0 & 60.9 \\

         \hline

        \multirow{7}{*}{ResNet-50} 
         & Bit Reduction~\cite{xu2017feature} & 2017  & 73.8 & 1.9 \\
         & Input Random~\cite{Xie2017Mitigating} & 2017 & 74.0 & 18.8 \\
         & Cheap-AT~\cite{wong2020fast} & 2020 & 55.6 & 26.2 \\
         & Jpeg~\cite{dziugaite2016study} & 2016 & 73.6 & 33.4 \\
         & Salman \textit{et al.}~\cite{salman2020adversarially} & 2020 & 64.0 & 35.0 \\
         & STL~\cite{sun2019adversarial} & 2019 & 68.3 & 50.2 \\
         & DISCO~\cite{disco_2022} &  2022 & 72.6 & 68.2 \\

        \hline

        \multirow{5}{*}{WRN-50-2}
         & Bit Reduction~\cite{xu2017feature} & 2017 & 75.1 & 5.0 \\
         & Input Random~\cite{Xie2017Mitigating}  & 2017 & 71.7 & 23.6 \\
         & Jpeg~\cite{dziugaite2016study} & 2016 & 75.4 & 24.9 \\
         & Salman \textit{et al.}~\cite{salman2020adversarially} & 2020 & 68.5 & 38.1 \\
         & DISCO~\cite{disco_2022} & 2022 & 75.1 & 69.5 \\

        \hline
    \end{tabular}
    \label{tab:comparison_defenses_AA_ImageNet}
\end{table}

The state-of-the-art results show that MNIST and CIFAR-10 datasets are already saturated. Other datasets should be further evaluated, namely: 1) \textbf{CIFAR-100 and ImageNet} since adversarial defenses do not achieve state-of-the-art clean accuracy (91\% and 95\%, respectively); 2) \textbf{GTSRB and SVHN}, depicting harder scenarios with greater variations of background, inclination, and luminosity; and 3) \textbf{Fashion-MNIST} that would allow better comprehension of which image properties influence DNNs performance (\textit{e.g.}, type of task, image shades, number of classes).

Most works present their results using accuracy as the evaluation metric and, more recently, evaluate their defenses on the Auto-Attack. Furthermore, the values given for $\epsilon$ in each dataset were standardized by recurrent use. However, there should be an effort to \textbf{develop a metric/process that quantifies the amount of perturbation added to the original image}. This would ease the expansion of adversarial attacks to other datasets that do not have a standardized $\epsilon$ value.

There has been a greater focus on the development of white-box attacks, which consider that the adversary has access to the network and training data, yet this is not feasible in real contexts, translating into the need of \textbf{focusing more on the development of black-box attacks}. A unique black-box set, \textbf{physical attacks, also require additional evaluation}, considering the properties of the real world and perturbations commonly found in it. Considering the increasing liberation of ML in the real world, end-users can partially control the training phase of DNNs, suggesting that \textbf{gray-box attacks will intensify} (access only to network or data).

The different network architectures are designed to increase the clean accuracy of DNNs in particular object recognition datasets, yet there should be \textbf{further evaluation on the impact of the different layers and their structure}. ViTs introduce a new paradigm in image analysis and are more robust against natural corruptions, suggesting that \textbf{building ViT inherently robust to adversarial examples} might be a possible solution.

DDPM are generative models that perform adversarial purification of images, but they can not be applied in real-time since they take up to dozens of seconds to create a single purified image. Therefore, an effort on \textbf{developing close to real-time adversarial purification strategies} is a viable strategy for future works.

\section{Conclusions}
\label{sec:conc}

DNNs are vulnerable to a set of inputs, denominated as adversarial examples, that drastically modify the output of the considered network and are constructed by adding a perturbation to the original image. This survey presents background concepts, such as adversary capacity and vector norms, essential to comprehend adversarial settings, providing a comparison with existing surveys in the literature. Adversarial attacks are organized based on the adversary knowledge, highlighting the emphasis of current works toward white box settings, and adversarial defenses are clustered into six domains, with most works exploring the adversarial training strategy. We also present the latest developments of adversarial settings in ViTs and describe the commonly used datasets, providing the state-of-the-art results in CIFAR-10, CIFAR-100, and ImageNet. Finally, we propose a set of open issues that can be explored for subsequent future works.

\section*{Acknowledgments}

This work was supported in part by the Portuguese FCT/MCTES through National Funds and co-funded by EU funds under Project UIDB/50008/2020; in part by the FCT Doctoral Grant 2020.09847.BD and Grant 2021.04905.BD;

\bibliographystyle{ieeetr}

\end{document}